\documentclass[letterpaper]{article} 
\usepackage{aaai24}  
\usepackage{times}  
\usepackage{helvet}  
\usepackage{courier}  
\usepackage[hyphens]{url}  
\usepackage{graphicx} 
\usepackage{natbib}  
\usepackage{caption} 
\title{Incomplete Contrastive Multi-View Clustering with
High-Confidence Guiding}

\author{
    Guoqing Chao,
    Yi Jiang,
    Dianhui Chu
}
\affiliations{

    Harbin Institute of Technology\\
    2 West Culture Road\\
    Weihai, Shandong 264209, China
}

\usepackage{bibentry}

\usepackage{tabularx}
\newcolumntype{Z}{>{\centering\let\newline\\\arraybackslash\hspace{0pt}}X}
\usepackage{bm}
\usepackage{multirow}
\usepackage{subfigure}
\usepackage[graphicx]{realboxes}
\usepackage{amsmath}
\usepackage{makecell}
\usepackage{pifont}
\usepackage{epstopdf}
\usepackage{epsfig}
\usepackage{algorithm}
\usepackage{algorithmicx}
\usepackage{algpseudocode}
\floatname{algorithm}{Algorithm} 
\usepackage{enumitem}

\begin{document}

\maketitle

\begin{abstract}
Incomplete multi-view clustering becomes an important research problem, since multi-view data with missing values are ubiquitous in real-world applications. Although great efforts have been made for incomplete multi-view clustering, there are still some challenges: 1) most existing methods didn't make full use of multi-view information to deal with missing values; 2) most methods just employ the consistent information within multi-view data but ignore the complementary information; 3) For the existing incomplete multi-view clustering methods, incomplete multi-view representation learning and clustering are treated as independent processes, which leads to performance gap. In this work, we proposed a novel Incomplete Contrastive Multi-View Clustering method with high-confidence guiding (ICMVC). Firstly, we proposed a multi-view consistency relation transfer plus graph convolutional network to tackle missing values problem. Secondly, instance-level attention fusion and high-confidence guiding are proposed to exploit the complementary information while instance-level contrastive learning for latent representation is designed to employ the consistent information. Thirdly, an end-to-end framework is proposed to integrate multi-view missing values handling, multi-view representation learning and clustering assignment for joint optimization. Experiments compared with state-of-the-art approaches demonstrated the effectiveness and superiority of our method. Our code is publicly available at https://github.com/liunian-Jay/ICMVC.
\end{abstract}

\begin{figure*}[t]
    \centering
    \includegraphics[height= 0.35\linewidth, width=0.99\linewidth]{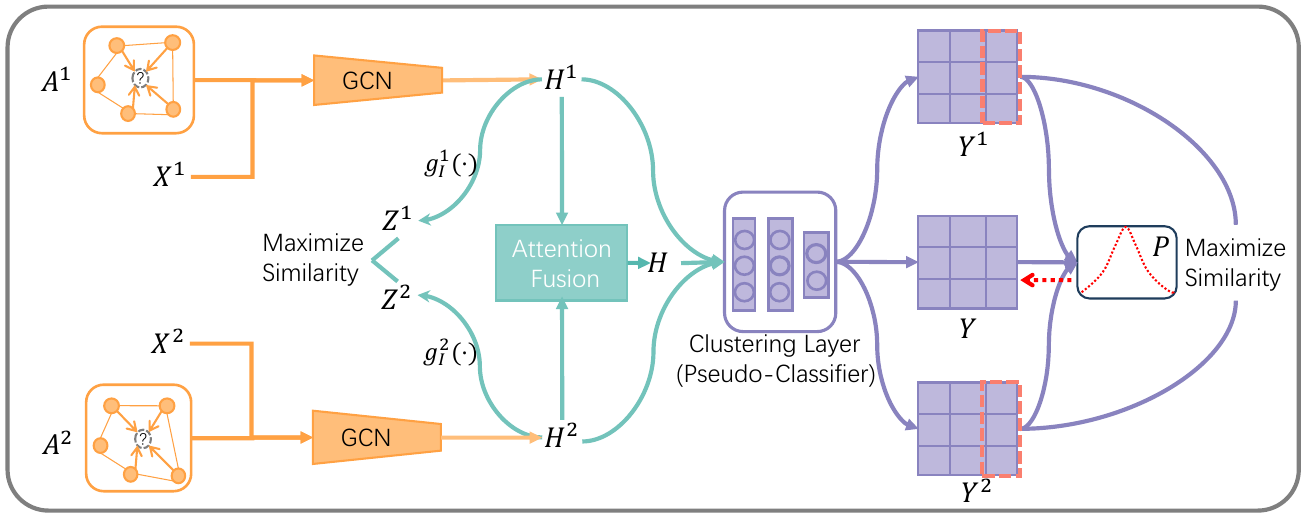}
    \caption{Schematic illustration of the proposed network architecture ICMVC (two views as an example).
    The three modules are represented with different colors. 
    Firstly, we handle missing data via multi-view consistency relation transfer and adopt GCN to encode incomplete multi-view data, where the representation of missing data is gradually learned with GCN message passing. Secondly, a common representation for multi-view fusion is then obtained through an attention module, while view-specific hidden representations are projected into the embedding space for instance-level comparative learning. 
    Finally, the clustering predictions are obtained through weight-sharing pseudo-classifier and target distributions are computed for high-confidence guidance.}
    \label{fig:model}
\end{figure*}

\section{Introduction}
\label{intro}

In practical applications, the majority of data can be considered as multi-view data~\cite{chao2021survey,yang2018multi} and exist missing values due to some  uncontrollable factors during data collection, transmission or storage. Incomplete Multi-View Clustering (IMVC) aims to learn a final clustering result based on the multi-view data with missing values.

Many IMVC algorithms have been proposed to solve the missing value problem within multi-view data~\cite{wen2022survey}.
The existing IMVC methods are generally based on non-negative matrix decomposition~\cite{hu2019doubly,chao2022incomplete}, subspace-based clustering methods~\cite{wang2022frobenius,chao2019analysis}, kernel learning~\cite{liu2019multiple,ye2017consensus}, graph spectral clustering~\cite{zhou2019consensus} or deep learning~\cite{lin2021completer,wen2021structural,ke2023clustering,ke2023disentangling}.
The popular IMVC methods PVC~\cite{li2014partial} and MIC~\cite{shao2015multiple} conduct multi-view clustering by ignoring the missing values, which leads to inferior performance. There are some other IMVC methods that adopted the missing value imputation methods to fill in the missing position in each view~\cite{wen2022survey} and then conducted multi-view clustering. These methods just use the relation between features within each view but ignore the relation between multiple views. 

Multi-view fusion plays a vital role in IMVC process, consistent information and complementary information should be exploited to finish this task. However,  most of current IMVC methods such as~\cite{xu2015multi} take good advantage of the consistent information within multi-view data but ignore the complementary information. Thus how to mine the complementary information within multi-view data and even make full use of both complementary and consistent information is worth further investigation.

Recently, deep IMVC methods have achieved great success due to their powerful representation learning capabilities.
Based on auto-encoder (AE), generative adversarial network (GAN), graph neural network (GNN), they first use existing data part to infer and fill in missing data, and then use traditional MVC methods for clustering.
For example, Completer~\cite{lin2021completer} trains the model with paired instances, and then completes the missing data by dual prediction while SDIMC-net~\cite{wen2021structural} uses GNN to tackle incomplete multi-view data problem. Due to the separate processes of missing data handling and multi-view clustering, their performance is not ideal.

To deal with the above problems, we proposed a novel end-to-end incomplete contrastive multi-view clustering method with high-confidence guiding. Firstly, we proposed a multi-view consistency relation transfer plus graph convolutional network (GCN) to handle missing values within multi-view data. Secondly, we designed an instance-level attention module and high-confidence guiding to exploit the complementary information while conducted instance-level contrastive learning for latent representation. Thirdly, we unified multi-view missing data handling, multi-view representation learning and clustering assignment for joint optimization to obtain the clustering result directly. The novelty existing in three stages may motivate more systematic investigation of IMVC. 

\begin{figure}[ht]
    \centering
    \includegraphics[width=1\linewidth]{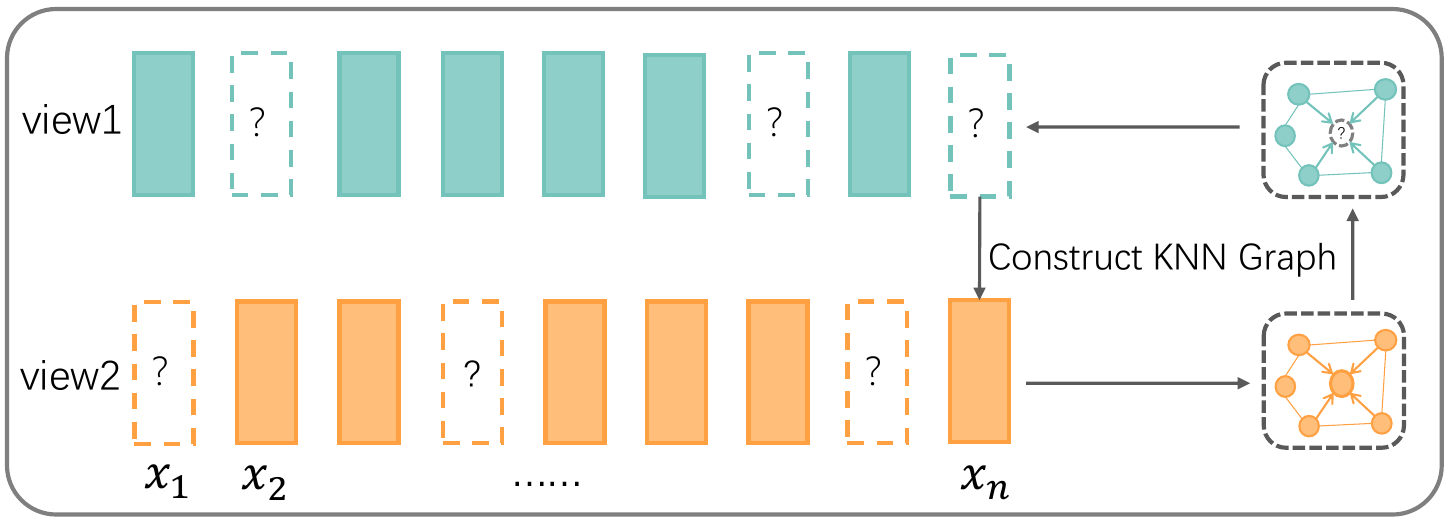}
    \caption{Illustration of the missing view case of multi-view data and how to obtain the adjacency matrix of the missing view via multi-view consistency relation transfer.}
    \label{fig:miss}
\end{figure}
\section{The proposed method}
\label{method}

\subsection{Notations}
We define the multi-view data including $N$ instances with $V$ views as $\bm{X}=\{ \bm{X}^1, \cdots,\bm{X}^v, \cdots,\bm{X}^V \}$, and $\bm{X}^v=\{\bm{x}_1^v, \cdots,\bm{x}_i^v, \cdots, \bm{x}_N^v \}\in \mathbf{R}^{N\times d_v}$ denotes the feature matrix of the $v$-th view, where $\bm{x}_i^v$ denotes   the $v$-th view feature of the $i$-th instance and $d_v$ is the feature dimension of the $v$-th view. 
The graph structure of the data is represented as an adjacency matrix $\bm{A}=\{\bm{A}^1, \cdots, \bm{A}^v, \cdots, \bm{A}^V\} $, and $\bm{A}_{ij}^v \in \mathbf{R}^{N\times N}$ indicates whether there is an edge between $\bm{x}_i^v$ and $\bm{x}_j^v$. 
For convenience, we will only discuss the two-view case, and the model can be easily extended to more than two views.

\subsection{Multi-View Missing Handling}
To build up the graph structure, we first compute the similarity matrix $\bm{S}^v \in \mathbf{R}^{n \times n}$ at each view according to the radial basis function: $\bm{S}_{ij}^v=e^{- \frac{{\left\| \bm{x}_i^v-\bm{x}_i^v \right\|}^2}{t} }$, where $\bm{S}_{ij}^v$ denotes the similarity between $\bm{x}_i$ and $\bm{x}_j$ from the $v$-th view. After that, we use the K-nearest neighbors to construct the graph structure $\bm{A}$ based on the similarity. 

\subsubsection{\textbf{Multi-View Consistency Relation Transfer}}
The missing view case of multi-view data and how to obtain the adjacency matrix of the missing view are shown in Figure \ref{fig:miss}. The dashed lines indicate the missing views, and thus the adjacency relation of the instance from the missing view cannot be obtained.
However, based on the consistency assumption between views in multi-view data,  the adjacency matrix obtained from the existing view of the same instance can be used as the adjacency matrix for the missing view. 
For instance, if the graph structure of instance $i$ is unavailable because its first view information is missing, we can transfer the graph structure of the second view to the first one, namely, $\bm{A}_i^1 = \bm{A}_i^2, i=1, 2,\cdots, N$, where $\bm{A}_i^1$ and $\bm{A}_i^2$ denote the $i$-th row of $\bm{A}^1$ and $\bm{A}^2$ respectively.  

\subsubsection{\textbf{GCN Missing Handling}}
After obtaining the adjacency matrix, we feed $\bm{A}^v$ and $\bm{X}^v$ with missing values into the GCN encoder. During the encoding process, benefiting from the message passing mechanism of GCN, the missing latent representations will be filled in. Specifically, this is estimated using the aggregated information of the neighbors of the missing instance. It should be noted that the missing values in $\bm X^v$ is not involved in the graph convolution operation due to zero in $A^v$ to be multiplied. The GCN encoder module will be explained in detail in the next subsection.
For extensions to the case of more than two views, the union or intersection of the graph structures of any existing views of an instance can be transferred to the missing view.

\subsection{Network Architecture}
As shown in Figure \ref{fig:model}, our model architecture consists of three modules: graph encoder module, multi-view fusion module, clustering module. High-confidence guidance is designed to exploit the complementary information within multi-view data. In the following subsections, we will describe them in detail. 

\subsubsection{\textbf{Graph Encoder Module}}
The graph auto-encoder captures not only the attribute information of the nodes, but also the graph structure information~\cite{scarselli2008graph,salehi2019graph,wang2019attributed}.
To learn the useful representation and deal with the missing data, we use stacked GCN layers as encoders.
Specifically, this module consists of $V$ view-specific encoders, which encode $V$ views respectively.
The mapping function for the $v$-th view can be expressed as 
$f_v(\bm{A}^v,\bm{X}^v;\theta^v) \rightarrow \bm{H}^v$, where $\bm{H}^v$ is the latent representation for the $v$-th view, and $\theta^v$ indicates the encoder parameter.
The graph convolution operation of the $m$-th layer is represented as
\begin{equation}
    \bm{H}_{(m)}^v=\phi (\Tilde{\bm{D}^v}^{-\frac{1}{2}}\Tilde{\bm{A}^v}\Tilde{\bm{D}^v}^{-\frac{1}{2}} \bm{H}_{(m-1)}^v \bm{W}_{(m)}^v),
\end{equation}
where $\Tilde{\bm{A}^v}=\bm{A}^v+\bm{I}$ and $\Tilde{\bm{D}^v}_{ii}=\sum_j \Tilde{\bm{A}^v}_{ij}$. $\bm{I}$ is the identity matrix and $\bm{W}_{(m)}^v$ indicates the trainable parameters in the $m$-th layer of the $v$-th view encoder and $\phi$ denotes the activation function, such as Relu~\cite{nair2010rectified}, etc. 
Note that here $\bm{H}^v_{(0)}=\bm{X}^v$, $\bm A^v$ is the complete graph structure obtained from the $v$-th view after multi-view consistency relation transfer, and $\bm X^v$ is the feature matrix with missing values from the $v$-th view.
To avoid learning a trivial solution, a skip connection~\cite{he2016deep} is introduced.
Mathematically, 
\begin{equation}
    \bm{H}^v_{(m)}=\bm{H}^v_{(m)}+\bm{H}^v_{(m-1)},
\end{equation}
where $\bm{H}^v_{(m-1)}$ is the feature representation learned in the $m-1$ layer.

\subsubsection{\textbf{Multi-View Fusion Module}}
Our multi-view fusion module consists of two parts: an instance-level attention module and a view alignment module utilizing instance-level contrastive learning.

Direct concatenation or average fusion of the multi-view features usually leads to sub-optimal clustering performance.
In order to make full use of the complementary information of different views, inspired by~\cite{vaswani2017attention}, we propose a  fine-grained instance-level attention approach for automatically perceiving view fusion weights, and its weights can be used to guide the modules in the network to reinforce each other.
Specifically, we calculate the multi-view common representation $\bm{H}$ by the following instance-level attention fusion formula:
\begin{equation}\label{eq:fusion}
    \bm{H} = \sum_{v=1}^V \bm {\bm \Lambda}^v \odot \bm{H}^v,
\end{equation}
where $\odot$ indicates the element-wise multiplication and $\bm {\Lambda}^v=[\bm {\lambda}^v,...,\bm {\lambda}^v] \in\mathbf{R}^{N \times d},v=1,...V$. $\bm {\lambda}^v \in \mathbf{R}^{N \times 1}$ is the attention coefficient computed by

\begin{equation}
\begin{split}
        \widetilde{\bm{H}} &= [\bm{H}^1, \bm{H}^2, ..., \bm{H}^V ],\\
        \widetilde{\bm{G}} &= f_u(\widetilde{\bm{H}}),\\
        [\bm{\lambda}^1,...,\bm{\lambda}^v,...,\bm{\lambda}^V] &= \mbox{sotfmax}(\mbox{sigmod}(\widetilde{\bm{G}})/\tau),\\
\end{split}
\end{equation}
where $f_u$ represents a nonlinear mapping like MLP, $d$ is the feature dimension of $\bm H$ and $[\cdot,..., \cdot]$  denotes the horizontal concatenation of a collection of matrices or vectors along row.
Sigmoid function before Softmax operation is a trick to avoid assigning a score close to one to a particular view~\cite{zhou2020end}.

With the instance-level attention module, complementary information within multi-view data is adequately utilized. In order to fully explore the consistency information, view representation distribution alignment is introduced as an auxiliary task of regularizing the encoder to keep the local geometry structure~\cite{trosten2021reconsidering}.
To achieve this goal, we employ instance-level contrastive learning.
Specifically, we project the potential representations of two views into the embedding space using two instance-level contrastive heads (ICH) consisting of two stacked nonlinear MLPs, $\bm{z}_i^v=g_I^v(\bm{h}_i^v)$.

In the embedding space, we use different views of the same instance to construct positive pairs and different instances to construct negative pairs.
We achieve view alignment by maximizing the similarity between positive instance pairs while minimizing the similarity between negative instance pairs.
To achieve this goal, the loss for the first view of the $i$th instance $\bm{x}_i^1$ is given as follows:
\begin{equation}
     \ell_i^1=-\log \dfrac{\exp{(s(\bm{z}_i^1,\bm{z}_i^2)/\tau_I})}{\sum_{j=1}^N[\exp{(s(\bm{z}_i^1,\bm{z}_j^1)/\tau_I)}+\exp{(s(\bm{z}_i^1,\bm{z}_j^2)/\tau_I)}]},
\end{equation}
where $\tau_I$ is the instance-level temperature parameter that controls the softness.
For the whole dataset, the loss function is given as follows:
\begin{equation}
    \mathcal{L}_{ins}=\frac{1}{2N} \sum_{i=1}^N(\ell_i^1 +\ell_i^2).
    \label{eq:ins_loss}
\end{equation}

\subsubsection{\textbf{Clustering Module}}
To directly obtain the cluster labels and achieve the end-to-end joint optimization, we design a pseudo-classifier that shares the weights among different views for clustering. Instances can be passed through a pseudo-classifier to obtain the soft assignments to clusters, ie. $\bm{Y}^v=g_C(\bm{H}^v)$, where $\bm{Y}^v \in \mathbf{R}^{N \times C}$ and $C$ is the numbers of the clusters.
For the convenience of expression, we denote the $j$-th column of $\bm{Y}^v$ as $\bm{y}_j^v$:
\begin{equation}
    y_j^v=
    \left[\begin{array}{c}
        \bm{Y}_{1j}^v\\
        ...\\
        \bm{Y}_{Nj}^v\\
    \end{array}\right],
\end{equation}
$\bm{y}_j^v$ collects the probability values of all the instances assigned to the $j$-th cluster, which can represent the corresponding cluster.
Same with ~\cite{huang2020deep}, we call it cluster-wise Assignment Statistics Vector (ASV).

ASVs from different clusters should be mutually exclusive (and ideally orthogonal), while ASVs from different views of the same cluster should be consistent. To achieve this goal, we extend contrastive clustering to multi-view data. 
Specifically, the contrastive loss for the $i$-th ASV in the first view is computed as follows:
\begin{equation}
    \hat{\ell}_j^1=-\log{\dfrac{\exp(s(\bm{y}_j^1,\bm{y}_j^2)/\tau_C)}{\sum_{k=1}^C[\exp(s(\bm{y}_j^1,\bm{y}_k^1)/\tau_C) + \exp(s(\bm{y}_j^1,\bm{y}_k^2)/\tau_C)]}},
\end{equation}
where $\tau_C$ is the cluster-level temperature parameter controlling the softness.
For the whole data, the cluster-level contrastive loss is represented as
\begin{equation}
     \mathcal{L}_{clu}=\frac{1}{2C} \sum_{j=1}^C(\hat{\ell}_j^1 +\hat{\ell}_j^2)-H(\bm{Y}^1)-H(\bm{Y}^2),
     \label{eq:clu_loss}
\end{equation}
where $H(\bm{Y}^v)=-\sum_{j=1}^{C}[P(\bm{y}^v_j)\log P(\bm{y}^v_j)]$ is the information entropy and $P(\bm{y}^v_j)=\frac{1}{N}\sum_{t=1}^N \bm{Y}^v_{tj}$.
Maximizing entropy is the strategy to avoid assigning all samples to the same cluster~\cite{hu2017learning,huang2020deep}.

\subsubsection{\textbf{High-Confidence Guiding}}
For unsupervised learning task, some works introduce self-supervised auxiliary objectives, such as DEC~\cite{xie2016unsupervised}.
For the probability assignment of the clustering module output, we hope to design an auxiliary objective to achieve three goals: 
1) Instances with consistency and high confidence can be self-supervised to further enhance their own representation learning.
2) Instances with one view of a high-confidence probability assignment and  other views of approximately uniform distribution assignment can be self-supervised to obtain a cluster assignment with a high-confidence probability.
3) For instances with all views of the approximately uniform distribution assignment, the auxiliary target can further blur them to position them at the cluster boundary.

Inspired by the above motivation, a target guidance that emphasizes the high-confidence instances is designed and introduced and it can take advantage of the complementary information. 
Specifically, each latent representation obtained from each view and their fused representation are fed into a weight-sharing pseudo-classifiers to obtain multiple cluster assignments. That is, $\bm{Y}^v=g_C(\bm{H}^v), v=1, 2$ and $\bm{Y}=g_C(\bm{H})$. $g_C$ indicates the weigh-sharing pseudo-classifier.
Based on these cluster assignment results, we can obtain the target assignment:
\begin{equation}
    \begin{split}
        \bm{q}_{ij} &= \mbox{max}\{\bm{Y}_{ij}^1, \bm{Y}_{ij}^2 ,\bm{Y}_{ij}\}, \\
    \end{split}
    \label{eq:tar1}
\end{equation}
which guarantees the high-confidence instance be emphasized. The confident target assignment with highest probability will be chosen as the target assignment. Highly confident assignments are enhanced and the instances at cluster boundary are further blurred by the following actions:
\begin{equation}
\begin{split}
    \bm{p}_{ij} &= \frac{\bm{q}_{ij}^2}{\sum_{j=1}^{k} \bm{q}_{ij}^2}, \\
\end{split}
\label{eq:tar2}
\end{equation}
which will be used as the auxiliary target distribution to guide the clustering results, and the optimization objective is formulated as
\begin{equation}
    \mathcal{L}_{hg} = KL(\bm{Y}\|\bm{P}) = \sum\limits_{i} \sum\limits_{j} \bm{p}_{ij}\log\frac{\bm{p}_{ij}}{\bm{y}_{ij}}.
    \label{eq:hg_loss}
\end{equation}

\renewcommand{\thealgorithm}{1} 
\begin{algorithm}
\caption{Optimization of the proposed ICMVC} 
    \label{algm}
    \begin{algorithmic}[1] 
        \Require Dataset $\bm X=\{\bm X^v,v=1,...,V\}$, numbers of cluster $C$, hyper-parameter K, temperature parameter $\tau_I,\tau_C$, total iteration numbers $epoches$
        \Ensure Clustering predictions
        \State Calculate the adjacency matrix $\bm{A}^1,\bm{A}^2$
        \State Transfer adjacency matrix to the missing views.
        \For{$epoch = 1 \to epoches$} 
            \State Compute the representations and soft assignment
            \Statex $\quad \bm{H}^1=f_1(\bm{X}^1,\bm{A}^1)$, $\bm{Z}^1=g_I(\bm{H}^1)$,$\bm{Y}^1=g_C(\bm{H}^1)$
            \Statex $\quad \bm{H}^2=f_2(\bm{X}^2,\bm{A}^2)$,  $\bm{Z}^2=g_I(\bm{H}^2)$,$\bm{Y}^2=g_C(\bm{H}^2)$
            \State Compute fused representations $\bm{H}$ by equation \eqref{eq:fusion}
            \State Compute the soft assignment $\bm{Y}=g_C(\bm{H})$
            \State Calculate the loss $\mathcal{L}_{ins}$ by equation \eqref{eq:ins_loss}
            \State Calculate the loss $\mathcal{L}_{clu}$ by equation \eqref{eq:clu_loss}
            \State Calculate the target distribution $\bm{P}$ by\eqref{eq:tar1}\eqref{eq:tar2}
            \State Calculate the loss $\mathcal{L}_{hg}$ by equation \eqref{eq:hg_loss}
            \State Calculate the overall loss $\mathcal{L}$ by equation \eqref{eq:overall_loss}
            \State Update through gradient descent to minimize $\mathcal{L}$
        \EndFor
        \State \Return $\bm Y$ //output
    \end{algorithmic}
\end{algorithm}

\subsection{Objective Function}
Our model is an end-to-end clustering method that does not require prior pre-training to complete missing views~\cite{lin2021completer}, nor does it require pre-training to initialize cluster centroids~\cite{xie2016unsupervised}, nor does it require additional k means clustering to obtain the final clustering results.
Thus we can simultaneously optimize the whole model, the total loss is represented as follows:
\begin{equation}
    \mathcal{L} = \mathcal{L}_{ins} + \mathcal{L}_{clu} + \mathcal{L}_{hg}.
    \label{eq:overall_loss}
\end{equation}
Although trade-off parameters can be added to balance the different losses, we just enforced the same weights and achieved good performance. Algorithm \ref{algm} summarizes the training process.

\section{Experiments}
\label{exp}

\begin{table*}[!ht]
\begin{center}
\resizebox{\textwidth}{!}{
    \begin{tabular}{clccccccccccccccc} \hline
    \multirow{2}{*}{\makecell{$\eta$}} & 
    \multirow{2}{*}{Method} & 
    \multicolumn{3}{c}{HandWritten} & 
    \multicolumn{3}{c}{Scene-15} & 
    \multicolumn{3}{c}{LandUse-21} & 
    \multicolumn{3}{c}{MSRC-V1} &
    \multicolumn{3}{c}{Noisy MNIST} \\
    &   
    & ACC & NMI & ARI 
    & ACC & NMI & ARI 
    & ACC & NMI & ARI 
    & ACC & NMI & ARI 
    & ACC & NMI & ARI \\ 
    \hline
    \multirow{9}{*}{0}
    &$BSV$
    &70.03	&69.99	&59.40
    &26.88	&25.38	&11.47	
    &19.70	&22.46	&6.40 	
    &63.43	&53.80	&44.46 
    &54.40  &48.49  &37.12\\
    &$Concat$    
    &73.20	&71.82	&$\underline{62.70}$
    &28.93	&28.11	&12.85
    &19.24	&24.11	&6.91 	
    &67.33	&58.72	&49.49
    &44.56  &46.64  &31.83\\
    &$PVC$ 
    &$-$ &$-$ &$-$
    &30.39	&$\underline{32.76}$ &15.67
    &$\underline{26.93}$ &$\underline{31.40}$	&$\underline{12.60}$
    &70.67	&63.19	&54.43
    &40.83  &35.50  &23.00\\
    &$MIC$   
    &$-$	&$-$	&$-$
    &33.09	&32.75	&16.51
    &22.95	&28.86	&9.41 
    &74.57	&65.29	&57.88
    &32.79  &31.91 &16.02\\
    &$DAIMC$  
    &$-$ &$-$ &$-$
    &29.08	&26.20	&12.47
    &24.33	&29.25	&10.44	
    &$\underline{78.76}$ &69.32	&62.73 
    &39.68  &37.11  &24.95\\
    &$Completer$       
    &$\underline{76.10}$ &$\underline{77.56}$ &61.94
    &$\underline{33.55}$ &31.84	&$\underline{18.18}$
    &25.32	&30.28	&10.32	
    &74.19	&62.55	&53.40 
    &81.82  &$\underline{82.44}$  &74.76\\
    &$DSIMVC$        
    &74.79	&71.48	&61.95	
    &19.95	&17.63	&7.95 	
    &19.70	&21.39	&6.67 
    &47.14	&39.53	&25.09 
    &$\underline{85.59}$  &80.10  &$\underline{76.51}$\\
    &$DIMVC$  
    &61.81	&63.48	&48.45
    &27.92	&23.45	&12.91
    &24.27	&31.32	&11.56
    &77.05	&$\underline{70.82}$ &$\underline{63.22}$ 
    &51.38  &52.66	&38.65\\ 
    &$Ours$
    &$\bm{84.95}$	&$\bm{83.62}$	&$\bm{78.01}$
    &$\bm{38.29}$	&$\bm{36.13}$	&$\bm{21.60}$
    &$\bm{27.76}$	&$\bm{31.57}$	&$\bm{14.50}$
    &$\bm{89.24}$	&$\bm{79.94}$	&$\bm{77.50}$ 
    &$\bm{97.94}$ 	&$\bm{94.57}$	&$\bm{95.51}$\\ 
    \hline
    \multirow{9}{*}{\makecell{0.3}} 
    &$BSV$    
    &66.01	&60.24	&46.60
    &26.47	&23.54	&9.30 
    &18.53	&20.89	&5.17 
    &60.19	&50.10	&37.58
    &49.99 	&46.35	&34.08\\
    &$Concat$    
    &66.51	&63.15	&50.52
    &26.78	&25.06	&10.42	
    &17.38	&21.36	&5.34 	
    &60.00	&49.22	&36.98 
    &44.96 	&45.30	&31.58\\
    &$PVC$  
    &$-$ &$-$ &$-$
    &29.51	&27.90	&13.62
    &$\underline{24.80}$	&28.04	&$\underline{10.64}$
    &50.10	&44.44	&29.06
    &45.32 	&35.19	&24.03\\
    &$MIC$       
    &$-$ &$-$ &$-$
    &28.32	&28.18	&12.28	
    &21.90	&25.79	&7.81 	
    &68.29	&59.45	&49.05
    &29.26 	&28.84  &12.76\\
    &$DAIMC$   
    &$-$ &$-$ &$-$
    &26.49	&22.25	&10.40	
    &22.84	&25.93	&8.47 	
    &$\underline{76.76}$ &$\underline{68.51}$	&$\underline{60.82}$ 
    &39.67 	&33.79	&22.60\\
    &$Completer$       
    &$\underline{76.61}$ &$\underline{77.84}$	&$\underline{63.16}$
    &$\underline{31.90}$ &$\underline{30.24}$ &$\underline{17.12}$	
    &24.09	&$\bm{31.24}$ &7.35 	
    &71.24	&61.93	&52.09
    &$\underline{79.96}$	&$\underline{80.08 }$	&$\underline{73.57}$\\
    &$DSIMVC$       
    &73.05	&69.64	&60.16
    &19.78	&17.44	&7.82 	
    &18.87	&19.91	&6.01 
    &48.10	&39.0	&24.52 
    &73.83  &67.25	&60.06\\ 
    &$DIMVC$
    &53.28	&50.29	 &36.71
    &28.26	&22.73	&12.82
    &23.70	&29.14	&10.05
    &73.43	&63.38	&55.24 
    &57.30 	&55.89	&44.11\\ 
    &$Ours$ 
    &$\bm{82.34}$	&$\bm{80.24}$	&$\bm{73.22}$
    &$\bm{36.20}$	&$\bm{34.21}$	&$\bm{19.69}	$
    &$\bm{26.94}$	&$\underline{29.67}$    &$\bm{12.92}$	
    &$\bm{87.72}$	&$\bm{76.97}$	&$\bm{74.03}$
    &$\bm{96.82}$   &$\bm{91.96}$ 	&$\bm{93.13}$\\ 
    \hline
    \end{tabular}
}
\caption{The clustering results of nine methods on five complete datasets and incomplete datasets with missing rate $\eta=0.3$,  the $1^{st}$ and $2^{nd}$ best results are indicated in bold and underlined, respectively. ``$-$" means the result is unavailable due to non-negative constraint violation.}
\label{table:evalution}
\end{center}
\end{table*}

\subsection{Experimental settings}
We implement ICMVC in PyTorch 1.12.1 and conduct all the experiments on Ubuntu 20.04 with NVIDIA 2080Ti GPU.
The Adam optimizer is adopted, and the learning rate is set to 0.001, the hyper-parameter K is set to 10.
The instance-level temperature parameter $\tau_I$ is fixed at 1.0, and the cluster-level parameter $\tau_C$ is fixed at 0.5.
We observe that it can fully converges after 500 epoches after training the network, thus 500 epoches is set to terminate.

To evaluate the proposed method on incomplete multi-view data, we randomly delete one view to construct missing data.
The missing rate $\eta$ is defined as $\eta = (n-m)/n$, where $m$ is the number of complete instances and $n$ is the number of all instances.

\subsection{Datasets and Metrics}
We used five commonly-used datasets in our experiments to evaluate our model. 
\textbf{HandWritten}:
It consists of 2000 samples of handwritten digits 0-9. 
We adopt Fourier coefficients of the character shapes and Karhunen-Love coefficients as two views.
\textbf{Scene-15}:
It consists of 4,485 images distributed in 15 scene categories with GIST and LBP features as two views.
\textbf{LandUse-21}:
It consists of 2100 satellite images from 21 categories with  two views: PHOG and LBP.
\textbf{MSRC-V1}:
It is an image dataset consisting of 210 images in seven categories, including trees, buildings, airplanes, cows, faces, cars, and bicycles, with GIST and HOG features as two views.
\textbf{Noisy MNIST}:
the original images are used as view 1, and the sampled intra-class images with Gaussian white noise are used as view 2, and we use its subset containing 10k samples in the experiments.


To evaluate the performance, three commonly-used clustering metrics: accuracy (ACC), normalized mutual information (NMI), and adjusted rand index (ARI) are adopted. The larger these metrics are, the better the clustering performance.

\subsection{Baselines and Experimental Results}
\begin{figure*}[tp]
	\centering
	\begin{minipage}{0.33\textwidth}
		\includegraphics[width=1\linewidth]{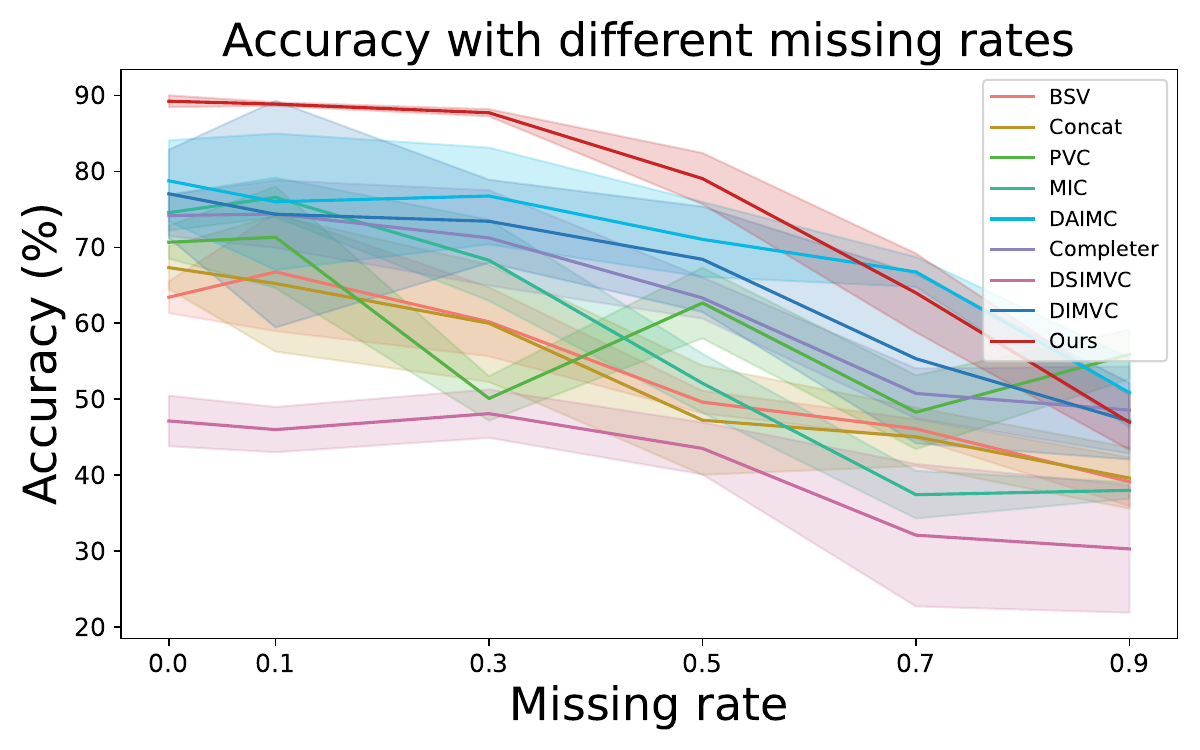}
	\end{minipage}
	\begin{minipage}{0.33\textwidth}
		\includegraphics[width=1\linewidth]{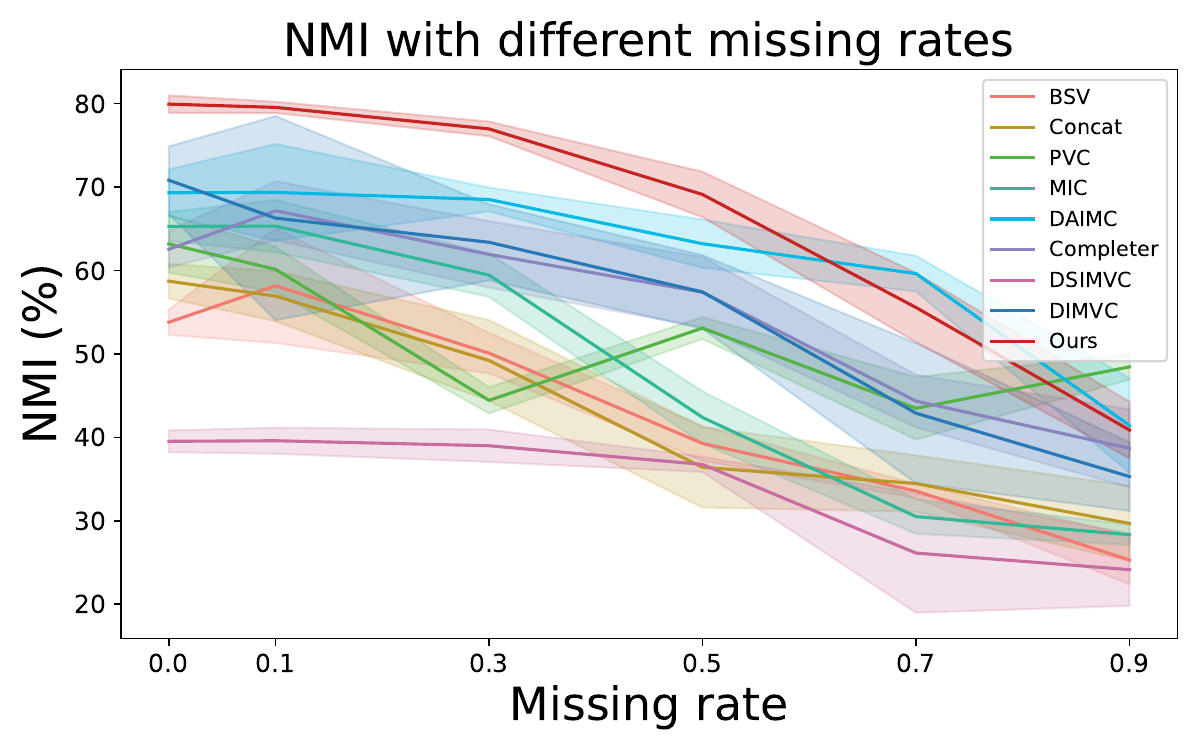}
	\end{minipage}
        \begin{minipage}{0.33\textwidth}
		\includegraphics[width=1\linewidth]{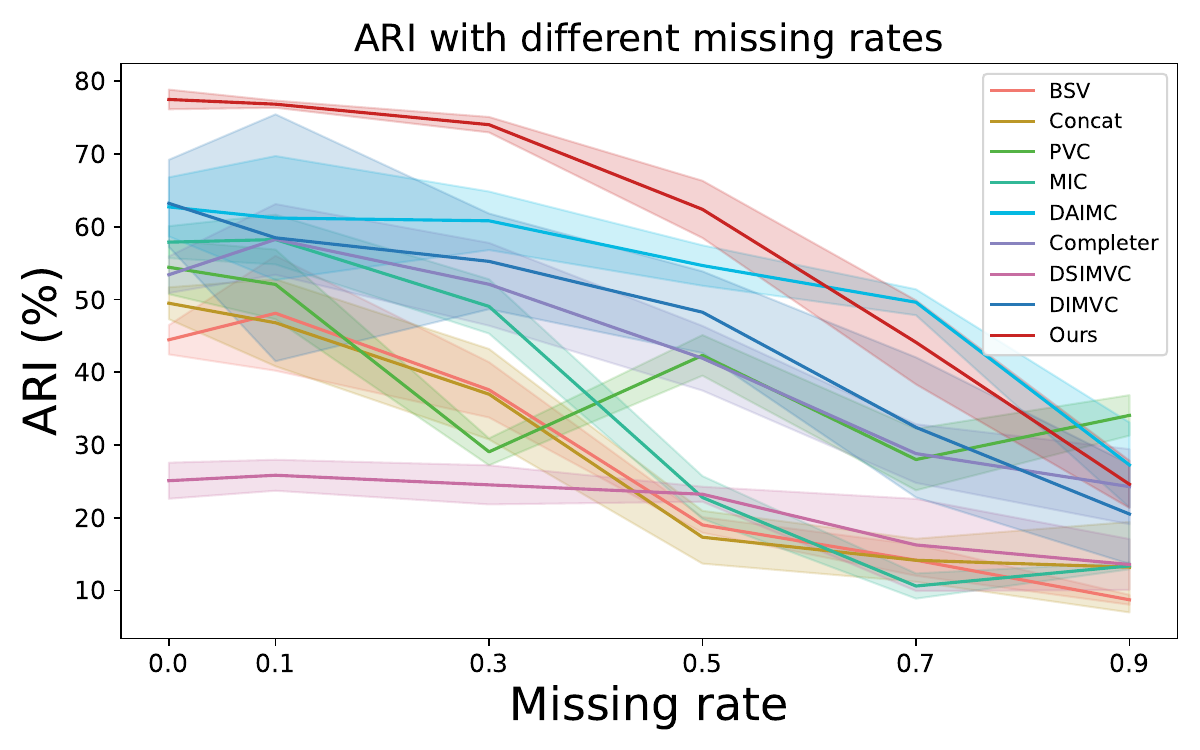}
	\end{minipage}
\caption{Error band plot of performance on MSRC-V1 as missing rate increases, with padding representing the standard deviation.}
\label{fig:error-band}
\end{figure*}
In order to verify the effectiveness and superiority of our method, two commonly-used single-view clustering methods and six state-of-the-art IMVC methods are used to compare with our proposed method ICMVC. These methods are listed as follows:
\begin{itemize}
\setlength{\itemsep}{0pt}
    \item \textbf{BSV}:    
    It performs k means clustering algorithm individually on each view and picks the best one as the final result.
    \item  \textbf{Concat}:
    It concatenates all the views and then performs the k means clustering algorithm to obtain the result.
    \item  \textbf{PVC}~\cite{li2014partial}: 
    It is a method based on NMF to learn the latent representations of incomplete multi-view data in subspaces.
    \item  \textbf{MIC}~\cite{shao2015multiple}: 
    It learns a latent representation for each view based on weighted NMF, and uses co-regularization to obtain a common representation.
    \item  \textbf{DAIMC}~\cite{hu2019doubly}: 
    It learns a consistent representation for all views using weighted NMF via the missing indicator matrix.
    \item  \textbf{Completer}~\cite{lin2021completer}: 
    It uses dual predictions to complete the latent representations for missing views, and then performs k means clustering algorithm on the latent representations.
    \item  \textbf{DSIMVC}~\cite{tang2022deep}: 
    It achieves safe IMVC by dynamically filling in missing views from the learned semantic neighborhoods.
    \item  \textbf{DIMVC}~\cite{xu2022deep}:
    It is an imputation-free and fusion-free IMVC method implemented via mining cluster complementarity.
\end{itemize}
Note that for the baseline methods BSV and Concat, we impute the missing values using the average feature value of that corresponding view of the existing instances.
For the other compared methods, the hyper-parameters are set to the recommended values in their original papers.

We set the missing rate $\eta=0.3$ to compare all the methods, and we also run them on the complete data.
To avoid randomness, we run all the methods five times, and then reported the average value of the clustering results, as shown in Table~\ref{table:evalution}. 

From Table~\ref{table:evalution}, we can find that compared with two single-view methods, our proposed method ICMVC performs better both in complete and incomplete settings. This demonstrates the effectiveness of multi-view fusion and missing value handling. Compared with other six state-of-the-art baselines, ICMVC outperforms them almost in every case.
In particular, on the MSRC-V1 dataset with missing rate 0.3, our method outperforms the state-of-the-art baselines by 10.96\%, 8.46\%, and 13.21\% in measures ACC, NMI, and ARI, respectively. This verifies the superiority of our proposed ICMVC. 

\subsection{Performance with Different Missing Rates}
In order to further verify the effectiveness of our model, we vary the missing rate from 0.1 to 0.9 with an interval 0.2 on MSRC-V1, and take the results on complete dataset as the starting point.

The error band diagram is shown in Figure \ref{fig:error-band}.
It can be clearly seen that at low missing rates, our method outperforms other baseline methods by far and the performance does not drop significantly as the missing rate increases.
However, the performance of our method degrades quickly when the missing rate is higher than 0.5. We guess that when the missing rate is too high, it is difficult to get the correct graph structure through the multi-view consistent relation transfer, thus the performance declines quickly.

Furthermore, we have observed that the standard deviation of the deep learning baseline methods is large, and we guess it may be easy to fall into a local optimal solution for deep learning methods due to some random reasons such as parameter initialization.
In comparison, the standard deviation of our method is relatively small, because our model has less dependence on the randomness of the parameter initialization and training process, and thus is robust.

\begin{table}[!ht]
\begin{center}
\resizebox{0.95\linewidth}{!}{
\begin{tabular}{c|ccc|ccc}
\hline
Missing Rate & $\mathcal{L}_{ins}$ &  $\mathcal{L}_{hg}$& $\mathcal{L}_{clu}$ &  ACC & NMI & ARI \\ 
\hline
\multirow{5}{*}{\makecell{0.3}} 
&\ding{51}&\ding{51}&\ding{51}&  \textbf{87.72} & \textbf{76.97} & \textbf{74.03} \\
&\ding{55}&\ding{51}&\ding{51}&  80.38 & 70.30 & 64.35 \\
&\ding{51}&\ding{55}&\ding{51}& 85.62 & 76.27 & 72.56 \\
&\ding{51}&\ding{55}&\ding{55}&  70.67 & 63.94 & 46.83 \\
\hline
\end{tabular}
}
\end{center}
\caption{Ablation study on MSRC-V1. The guidance module is only valid if the clustering module exists, and if $\mathcal{L}_{clu}$ is removed we use k means to perform clustering.}
\label{table:ablation}
\end{table}

\begin{figure*}[!ht]
\vspace{0pt}
	\centering
	\begin{minipage}{0.31\textwidth}
		\includegraphics[width=1\linewidth]{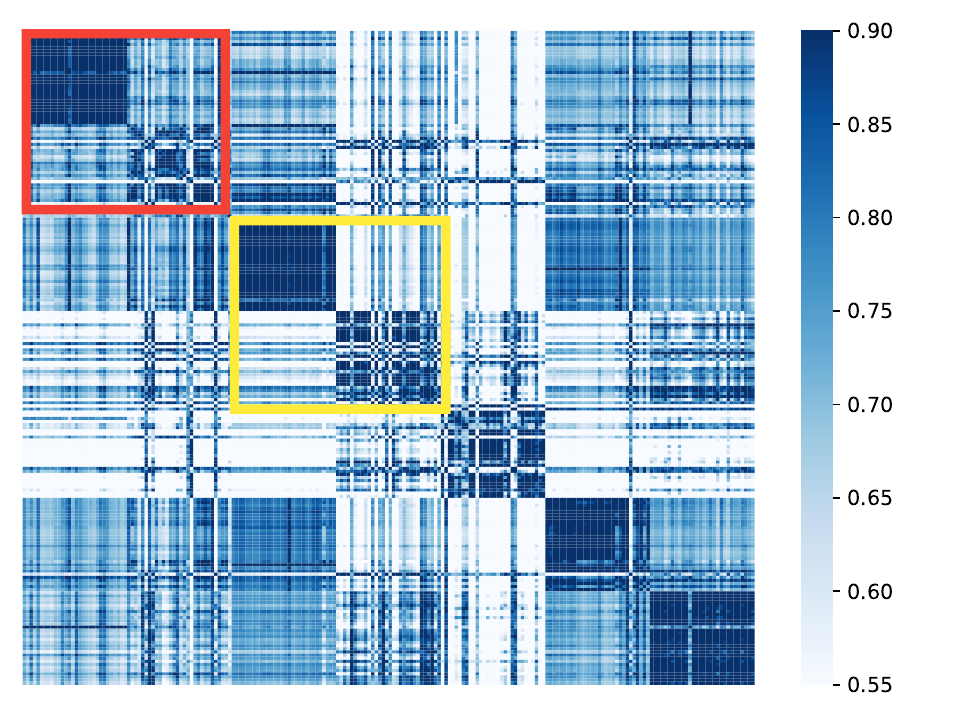}
	\end{minipage}
	\begin{minipage}{0.31\textwidth}
		\includegraphics[width=1\linewidth]{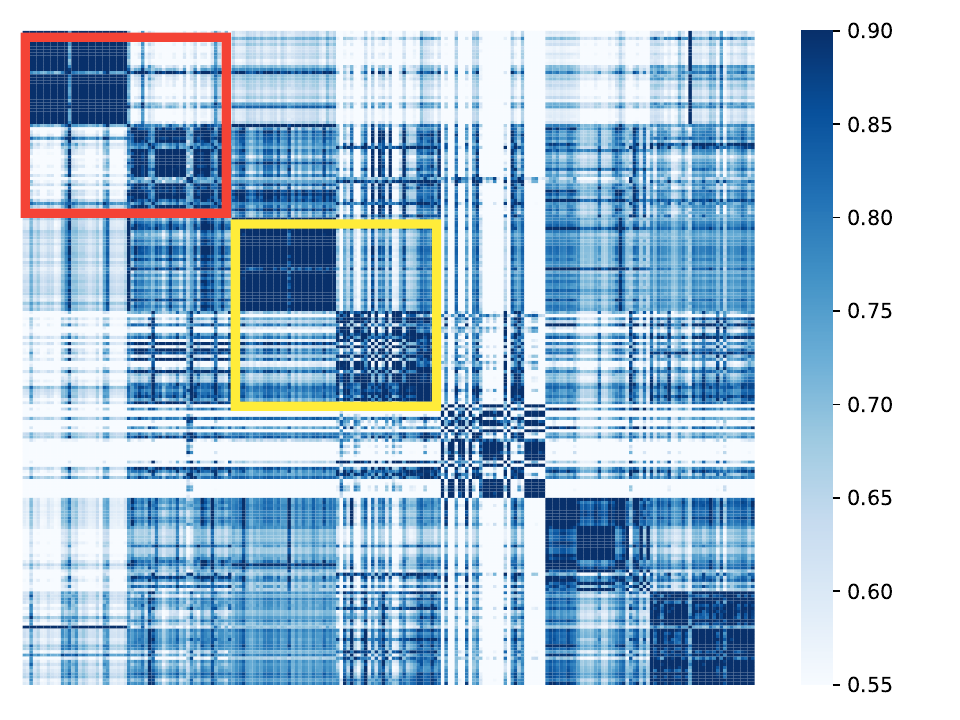}
	\end{minipage}
        \begin{minipage}{0.31\textwidth}
		\includegraphics[width=1\linewidth]{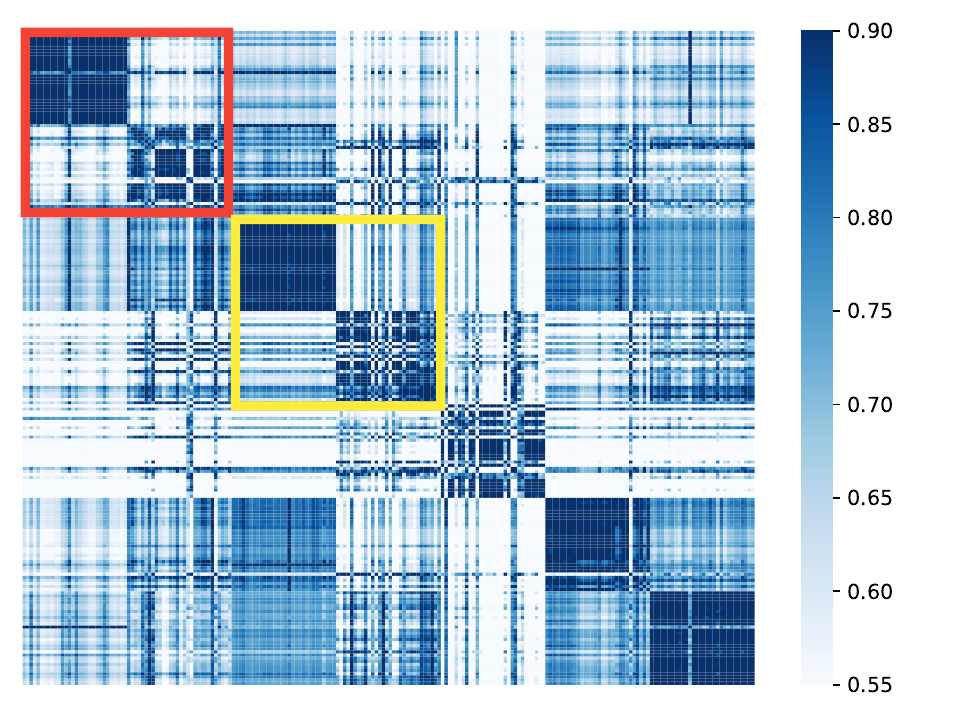}
	\end{minipage}
        \begin{tabularx}{0.95\linewidth}{ZZZZ}
        View one & View two & Fused view
        \end{tabularx}
\caption{Visualization of the similarity matrix obtained from each view and fused view of MSRC-V1 with missing rate 0.3.}
\label{fig:vis}
\end{figure*}

\begin{figure}[bp]
	\centering
	\begin{minipage}{0.494\linewidth}
		\includegraphics[width=1\linewidth]{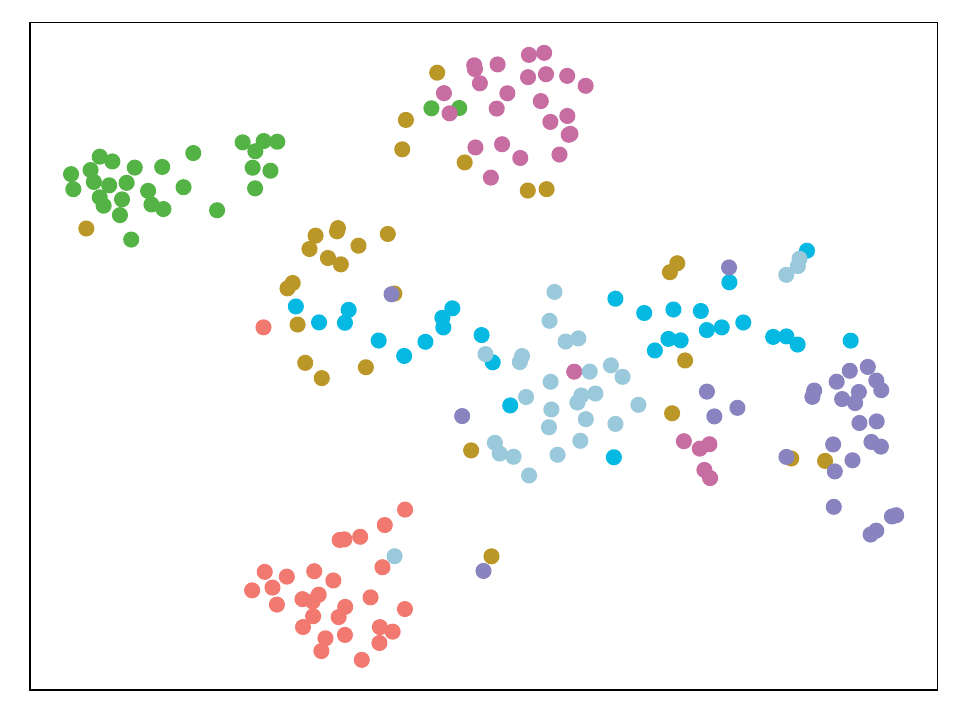}
	\end{minipage}
        \begin{minipage}{0.495\linewidth}
		\includegraphics[width=1\linewidth]{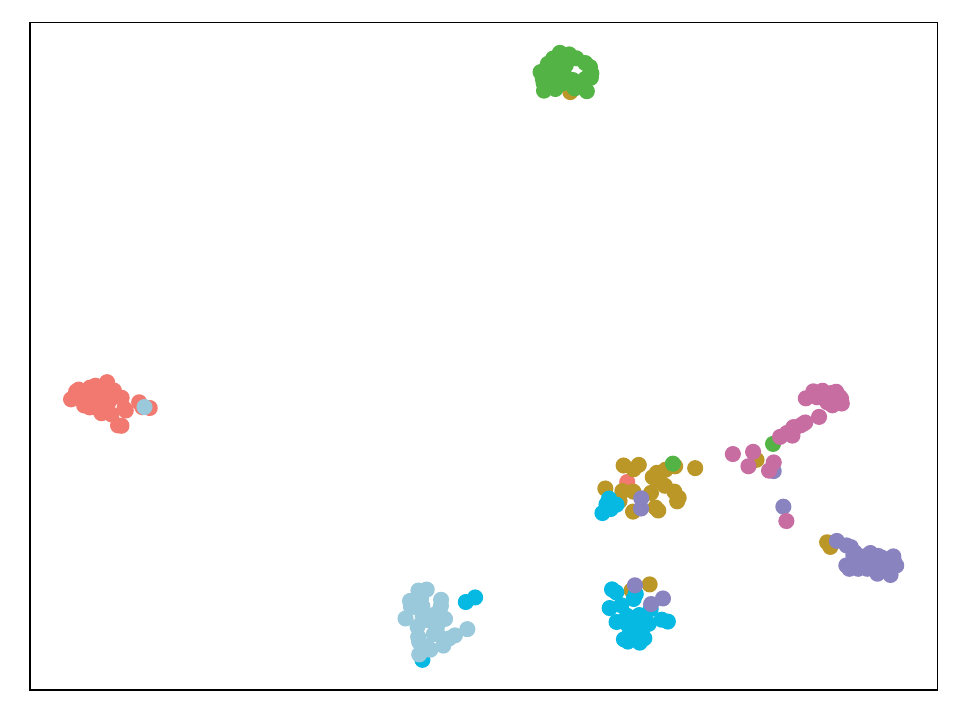}
	\end{minipage}
        \begin{tabularx}{0.95\linewidth}{ZZ}
         DAIMC & ICMVC(ours)
        \end{tabularx}
\caption{t-SNE visualization of the latent representations learned by DAIMC and ICMVC on MSRC-V1 with $\eta=0.3$.}
\label{fig:vis-methods}
\end{figure}

\subsection{Ablation Study}
To investigate and explore the effectiveness of each module of our model, we performed the ablation study experiments on MSRC-V1 with missing rate $\eta=0.3$.
Specifically, we separate each module and then retrain each one to conduct experiments. It should be noted that since we use a pseudo-classifier to predict cluster labels, if $\mathcal{L}_{clu}$ is removed, the clustering results will no longer be valid, and target guidance will lose its value. 
Thus we remove these two modules at the same time, and then perform k means clustering on the latent representation. 
The experimental results are shown in Table \ref{table:ablation}. 
It can be clearly seen that each module plays an important role.

\subsection{Visualization Analysis}
To compare with other competitive methods, we use t-SNE to visualize the latent representations of the other top method DAIMC, as shown in Figure \ref{fig:vis-methods}. We can find that the latent representation obtained from DAIMC looks good, but there are still some overlaps between different clusters. In comparison, our method ICMVC performs better, which maybe because it makes full use of the consistent and complementary information within multi-view data.

Based on the learned latent representations $\bm{H}, \bm {H^1}, \bm {H^2}$, we compute the similarity matrix using cosine distance and visualize them in Figure \ref{fig:vis}. From Figure \ref{fig:vis}, we can clearly see the block structure of the matrix. That is, the instance similarity within the same cluster is high, and the instance similarity between different clusters is low. We also observed that the similarity matrix obtained from fused view has better structure than that from each view. For example, there are some samples that are difficult to distinguish in the single view, such as the red box in view 1 and the yellow box in view 2. 
Instance-level attention fusion and high-confidence guiding take full advantage of this complementarity, and the resulting representations are all discriminative, as seen from the fused view.

\begin{figure}[!ht]
    \centering
    \includegraphics[width=0.80\linewidth]{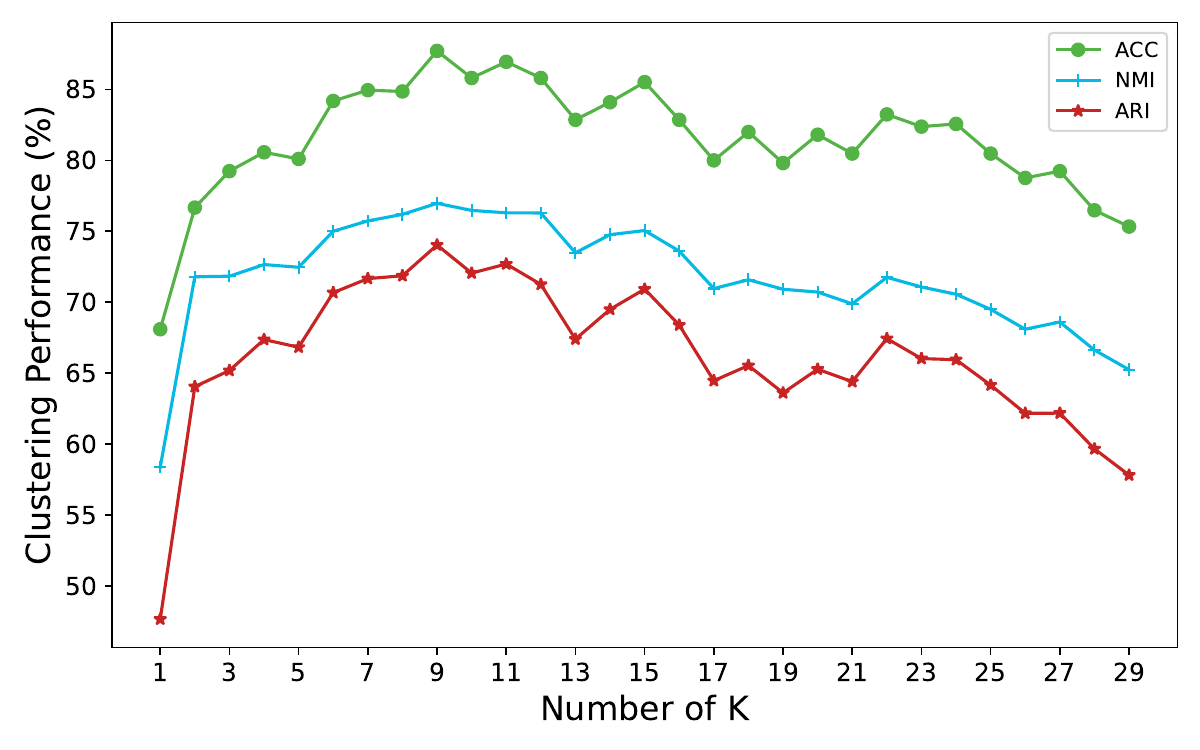}
    \caption{Illustration of the sensitivity analysis of K on MSRC-V1, x-axis represents the number of K, and y-axis represents the clustering performance.}
    \label{fig:k-num}
\end{figure}

\begin{figure}[!ht]
    \centering
    \includegraphics[width=0.83\linewidth]{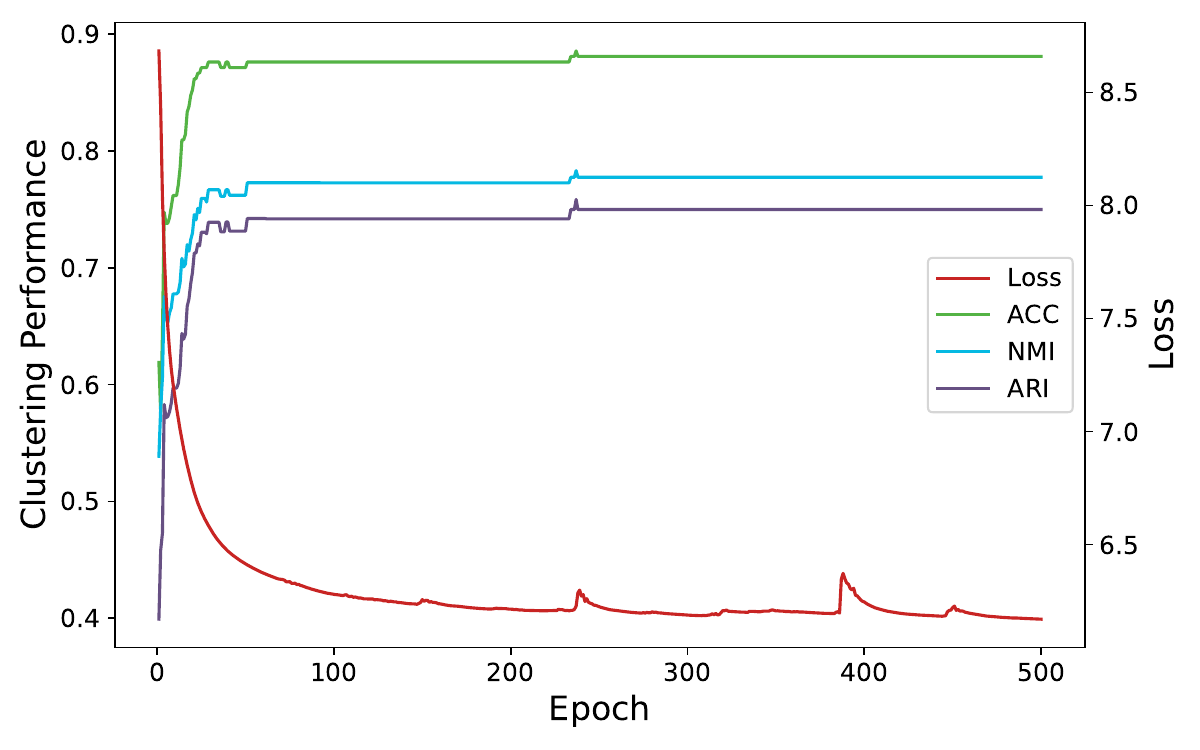}
    \caption{Illustration of the convergence analysis, x-axis represents number of training epoches, and left and right x-axes represent clustering performance and loss, respectively.}
    \label{fig:loss}
\end{figure}

\subsection{Parameter Sensitivity Analysis}
In this study, there is only one hyper-parameter K when constructing the graph structure in multi-view missing data handling stage. We conduct parameter sensitivity analysis for K on MSRC-V1 with missing rate 0.3, as shown in Figure \ref{fig:k-num}.
Since there are 30 instances for each cluster on MSRC-V1, we set K from 1 to 30 as shown in x axis, and we demonstrate the clustering performance by varying K, as shown in y axis in Figure~\ref{fig:k-num}. It is easy to see that when K is too small or too large, the clustering performance is bad. This may be because when K is small, too few neighbors cannot capture the precise structural information and when K is too large, a wrong adjacency relation will be established. We think the K value will be in a reasonable range if it take values not too small and not more than half of the number of instances in a cluster.
Within the reasonable range, the model performance fluctuates slightly with the K value, and the overall performance is stable, thus our model is insensitive to the K value within this suitable range.

\subsection{Convergence Analysis}
We show the convergence of ICMVC on MSRC-V1 with a missing rate 0.3 in Figure \ref{fig:loss}. As shown in Figure \ref{fig:loss}, we can clearly observe that in the first 100 epoches, the loss drops rapidly and ACC, NMI, and ARI rise steadily. After that, as the number of epoches increase, the loss decreases slowly with fluctuations, while ACC, NMI, and ARI rise steadily and eventually converge.

\section{Conclusion}
In this paper, we propose an end-to-end framework for incomplete multi-view clustering with high-confidence guiding. 
We designed a multi-view consistency relation transfer plus GCN to handle the missing values and make full use of the complementary information by designing instance-level attention module and high-confidence guiding. 
Moreover, our method unites multi-view missing data handling, multi-view representation learning and clustering into a unified framework to obtain the clustering results without pre-training or post-processing. Experiments on five datasets verified the effectiveness and superiority of the proposed method ICMVC. For future work, this method can be easily extended to more than two views, but the computation complexity increase dramatically with view number increasing, thus we will continue to explore more efficient solutions. In current work, we just treat three losses equally and get excellent performance, next we can consider enforcing different weights and learn the wights adaptively.
\section{Acknowledgments}
I commit to an author presenting the paper in-person at AAAI-24. If exceptional circumstances arise that prevent the authors from presenting the paper in person, I will notify AAAI immediately. I acknowledge that if I do not notify AAAI about exceptional circumstances, then the paper will not be included in the proceedings.

This work is supported in part by the National Natural Science Foundation of China (No. 62276079), Young Teacher Development Fund of Harbin Institute of Technology IDGA10002071, Research and Innovation Foundation of Harbin Institute of Technology IDGAZMZ00210325, Key Research and Development Plan of Shandong Province 2021SFGC0104 and the Special Funding Program of Shandong Taishan Scholars Project.

\bibliography{aaai24}

\clearpage
\appendix
\section{Datasets}
We adopt five widely used public datasets for evaluation, including HandWritten, Scene-15, LandUse-21, MSRC-v1 and Noisy MNIST. The details of these datasets are shown in Table \ref{table:dataset}.
\begin{table}[!ht]
\begin{center}
\resizebox{0.95\linewidth}{!}{
\begin{tabular}{ccccc}
\hline
Dataset & $\#$Samples & $\#$Categories & $\#$Views & $\#$Dims\\ 
\hline
HandWritten & 2000 & 10 & 2 & 76,64 \\
Scene-15    & 4485 & 15 & 2 & 59,40 \\
LandUse-21  & 2100 & 21 & 2 & 59,40 \\
MSRC-V1     & 210  & 7  & 2 & 576,512 \\ 
Noisy MNIST & 10000& 10 & 2 & 784,784 \\ 
\hline
\end{tabular}
}
\end{center}
\caption{The statistics of five datasets.}
\label{table:dataset}
\end{table}

\section{Visualization}
\begin{figure}[htp]
\centering
\begin{tabular}{cc}
	\includegraphics[width=0.45\linewidth]{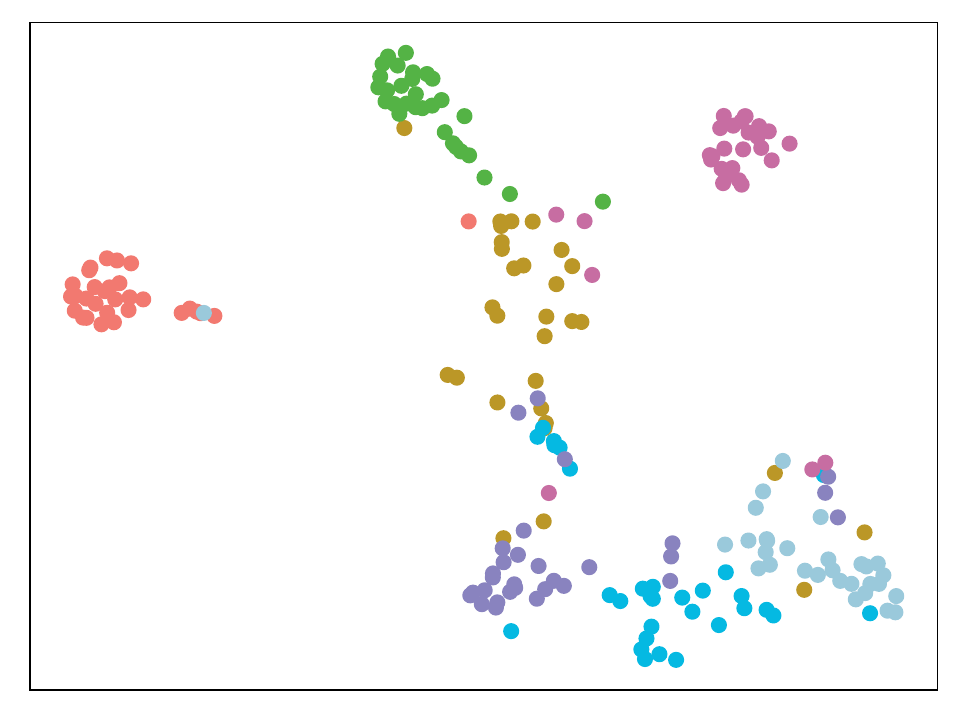} &
    \includegraphics[width=0.45\linewidth]{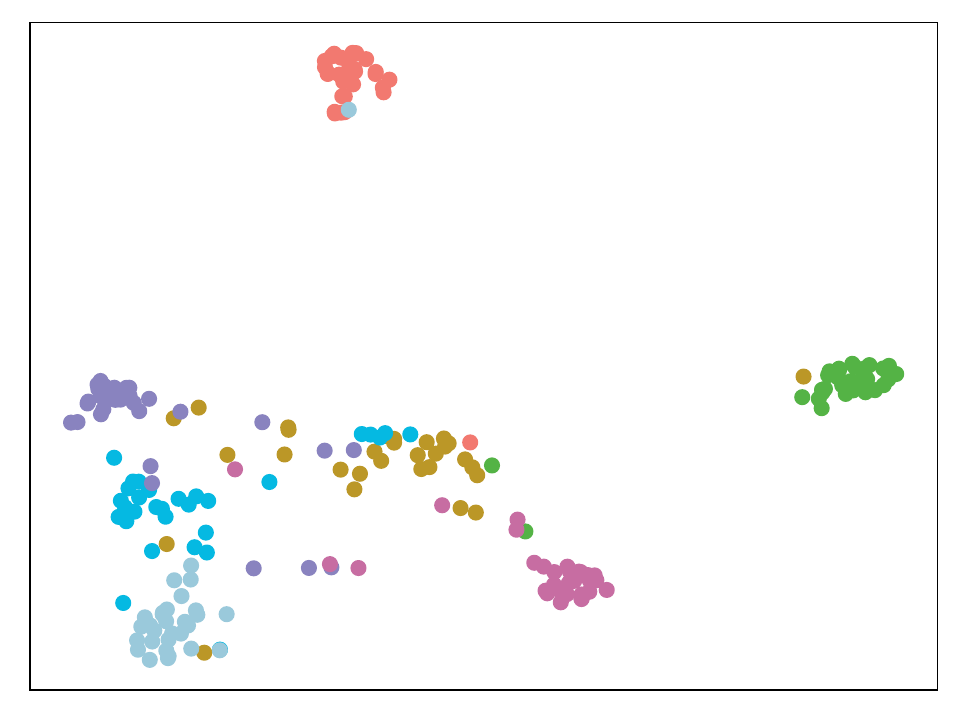}  \\
        1 epoch & 25 epoch \\
    \includegraphics[width=0.45\linewidth]{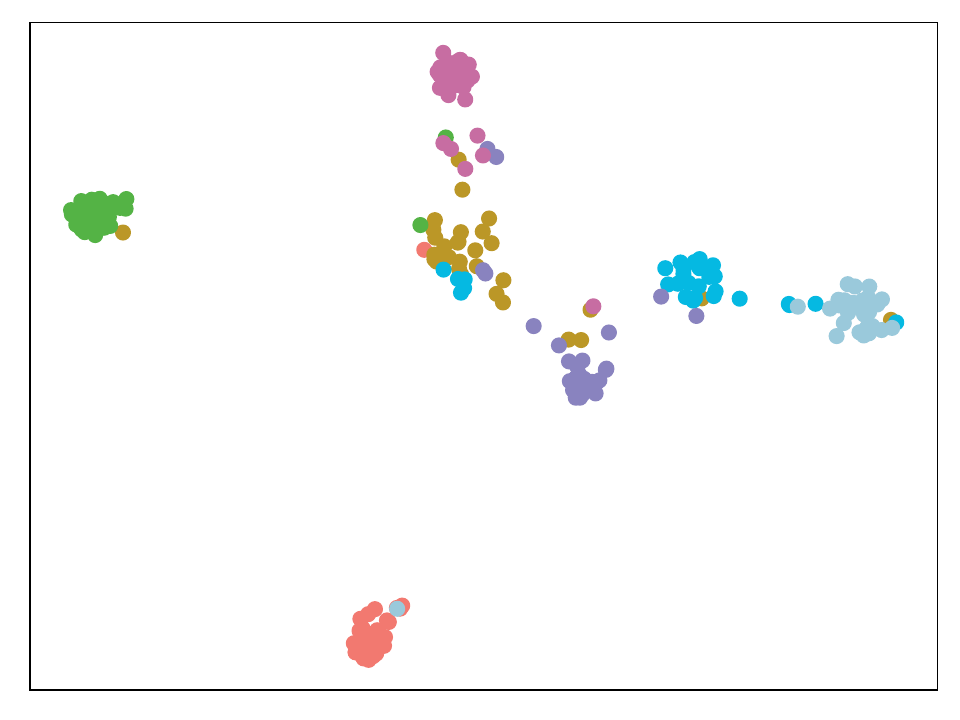} &
    \includegraphics[width=0.45\linewidth]{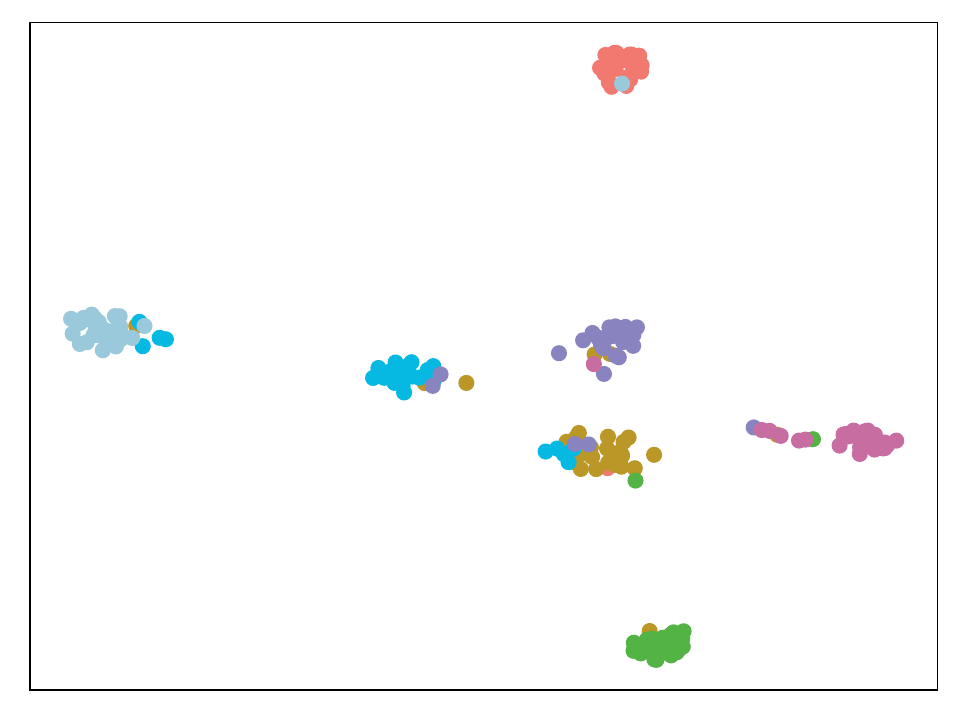}  \\
        50 epoch & 100 epoch \\
\end{tabular}
    \captionof{figure}{T-SNE visualization of the latent representations obtained in different training epochs on MSRC-V1 with missing rate 0.3. }
\label{figure:epoch-vis}
\end{figure}

In order to show the effectiveness of the algorithm more clearly, we supplement the change of the latent representation as the number of iterations increases. As shown in Figure~\ref{figure:epoch-vis}, as the training progresses, the learned fusion representation is more and more conducive to clustering. Finally a clear clustering structure emerges, achieving separation between clusters and compactness within clusters.

\begin{figure}[ht]
\centering
\begin{tabular}{cc}
        \includegraphics[width=0.45\linewidth]{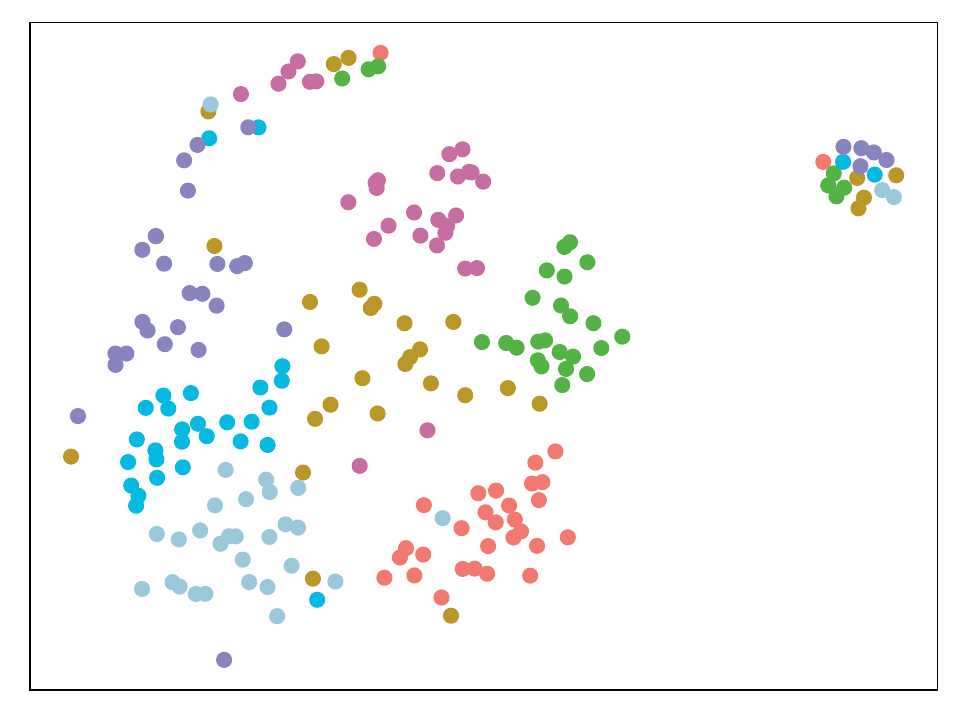} &
        \includegraphics[width=0.45\linewidth]{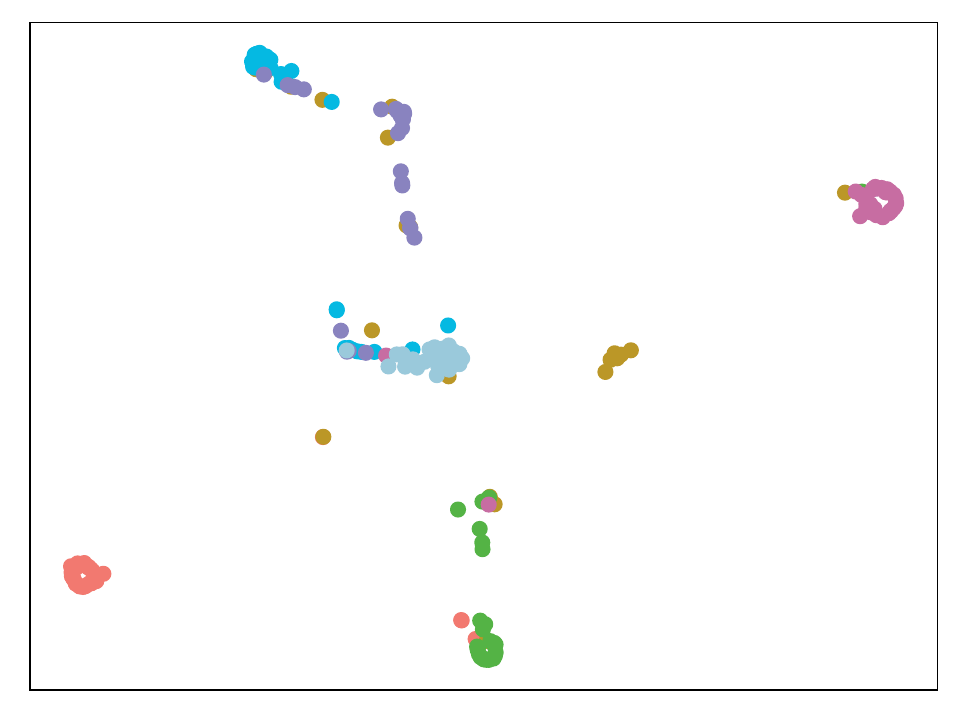}  \\
        Concat & Completer-15 \\
        \includegraphics[width=0.45\linewidth]{v-DAIMC.pdf} &
        \includegraphics[width=0.45\linewidth]{v-ours.pdf}  \\
        DAIMC-21 & ICMVC(ours) \\
\end{tabular}
    \captionof{figure}{T-SNE visualization of the latent representations obtained by different methods on MSRC-V1 with missing rate 0.3. }
\label{figure:method-vis}
\end{figure}

Moreover, we also show here the visualization results compared with more other methods. The visual structure of the latent representation is shown in Figure~\ref{figure:method-vis}. We can see that the representation learned by our method has a clearer cluster structure, which is more conducive to clustering.

\section{Detailed experimental results}

\subsection{Experimental setup}
It should be noted that we unify the settings of the experiments since we find that when we run multiple experiments to calculate the standard deviation, different methods used different experimental settings. Specifically, some methods (such as COMPLETER, etc.) adopt the strategy of fixing the randomness of the model and randomly missing data, while other methods adopt the strategy of initializing the model differently each time and inputting the same batch of missing data. For the sake of fairness, we adopt the latter, and also reset all compared methods to the latter settings for experiments.
\subsection{Experimental results}

\begin{table*}[!pt]
\resizebox{0.99\textwidth}{!}{
\begin{tabular}{c|l|ccc|ccc} \hline
\multirow{2}{*}{\makecell{Dataset}} & 
\multirow{2}{*}{Method} & \multicolumn{3}{c|}{Complete} & \multicolumn{3}{c}{Incomplete (30$\%$ missing)}\\
 & & ACC & NMI & ARI  & ACC & NMI & ARI \\ 
\hline
\multirow{9}{*}{MSRC-V1} 
&$BSV$  	
& 63.43 $\pm$ 2.09 & 53.80 $\pm$ 1.52 & 44.46 $\pm$ 2.01
& 60.19 $\pm$ 4.54 & 50.10 $\pm$ 2.43 & 37.58 $\pm$ 3.80 \\
&$Concat$    
& 67.33 $\pm$ 2.81 & 58.72 $\pm$ 2.06 & 49.49 $\pm$ 2.19
& 60.00 $\pm$ 7.72 & 49.22 $\pm$ 4.84 & 36.98 $\pm$ 6.22 \\
&$PVC$ \cite{li2014partial}   
& 70.67 $\pm$ 2.15 & 63.19 $\pm$ 3.43 & 54.43 $\pm$ 3.69
& 50.10 $\pm$ 2.92 & 44.44 $\pm$ 1.58 & 29.06 $\pm$ 1.85 \\
&$MIC$ \cite{shao2015multiple}       
& 74.57 $\pm$ 2.37 & 65.29 $\pm$ 1.73 & 57.88 $\pm$ 2.20
& 68.29 $\pm$ 5.34 & 59.45 $\pm$ 2.60 & 49.05 $\pm$ 3.73 \\
&$DAIMC$ \cite{hu2019doubly} 
& $\underline{78.76 \pm 5.33}$ & 69.32 $\pm$ 2.81 & 62.73 $\pm$ 4.06
& $\underline{76.76 \pm 6.37}$ & $\underline{68.51 \pm 1.43}$ & $\underline{60.82 \pm 4.02}$ \\
&$Completer$ \cite{lin2021completer}        
& 74.19 $\pm$ 2.70 & 62.55 $\pm$ 2.21  & 53.40 $\pm$ 2.52
& 71.24 $\pm$ 6.27 & 61.93 $\pm$ 3.98 & 52.09 $\pm$ 5.69 \\
&$DSIMVC$ \cite{tang2022deep}     
& 47.14 $\pm$ 3.33 & 39.53 $\pm$ 1.31 & 25.09 $\pm$ 2.48
& 48.10 $\pm$ 3.19 & 39.02 $\pm$ 1.95 & 24.52 $\pm$ 2.69 \\
&$DIMVC$  \cite{xu2022deep} 
& 77.05 $\pm$ 5.80 & $\underline{70.82 \pm 4.06}$ & $\underline{63.22 \pm 5.98}$
& 73.43 $\pm$ 5.47 & 63.38 $\pm$ 4.56 & 55.24 $\pm$ 6.57 \\
&$ICMVC (Ours)$ 
& $\bm{89.24 \pm 0.77}$ & $\bm{79.94 \pm 1.05}$  & $\bm{77.50 \pm 1.35}$
& $\bm{87.72 \pm 0.47}$ & $\bm{76.97 \pm 0.88}$ & $\bm{74.03 \pm 1.06}$ \\ 
\hline

\multirow{9}{*}{Scene-15} 
&$BSV$  	
 & 26.88 $\pm$ 1.22 & 25.38 $\pm$ 1.18 & 11.47 $\pm$ 0.52
& 26.47 $\pm$ 1.18 & 23.54 $\pm$ 0.37 & 9.30 $\pm$ 0.55 \\
&$Concat$    
& 28.93 $\pm$ 0.50 & 28.11 $\pm$ 0.99 & 12.85 $\pm$ 0.28
& 26.78 $\pm$ 0.60 & 25.06 $\pm$ 0.32 & 10.42 $\pm$ 0.24 \\
&$PVC$ \cite{li2014partial}   
& 30.39 $\pm$ 0.70 & $\underline{32.76 \pm 0.68}$ & 15.67 $\pm$ 0.58
& 29.51 $\pm$ 0.63 & 27.90 $\pm$ 0.14 & 13.62 $\pm$ 0.35 \\
&$MIC$ \cite{shao2015multiple}       
& 33.09 $\pm$ 1.10 & 32.75 $\pm$ 1.41 & 16.51 $\pm$ 0.83
& 28.32 $\pm$ 0.85 & 28.18 $\pm$ 0.97 & 12.28 $\pm$ 0.88 \\
&$DAIMC$ \cite{hu2019doubly} 
& 29.08 $\pm$ 1.13 & 26.20 $\pm$ 0.90 & 12.47 $\pm$ 0.63
& 26.49 $\pm$ 1.29 & 22.25 $\pm$ 0.72 & 10.40 $\pm$ 0.36 \\
&$Completer$ \cite{lin2021completer}        
& $\underline{33.55 \pm 1.55}$ & 31.84 $\pm$ 1.09 & $\underline{18.18 \pm 0.79}$
& $\underline{31.90 \pm 1.31}$ & $\underline{30.24 \pm 0.51}$ & $\underline{17.12 \pm 0.74}$ \\
&$DSIMVC$ \cite{tang2022deep}     
&  19.95 $\pm$ 0.65 & 17.63 $\pm$ 0.15 & 7.95 $\pm$ 0.23
& 19.78 $\pm$ 0.13 & 17.44 $\pm$ 0.12 & 7.82 $\pm$ 0.17 \\
&$DIMVC$  \cite{xu2022deep} 
& 27.92 $\pm$ 1.57 & 23.45 $\pm$ 0.89 & 12.91 $\pm$ 0.94
& 28.26 $\pm$ 0.93 & 22.73 $\pm$ 0.54 & 12.82 $\pm$ 1.23 \\
&$ICMVC (Ours)$ 
& $\bm{38.29 \pm 1.75}$ & $\bm{36.13 \pm 0.49}$  & $\bm{21.60 \pm 0.94}$
& $\bm{36.20 \pm 1.47}$ & $\bm{34.21 \pm 0.52}$  & $\bm{19.69 \pm 0.86}$ \\ 
\hline
\multirow{9}{*}{LandUse-21} 
&$BSV$  	
& 19.70 $\pm$ 1.34 & 22.46 $\pm$ 0.93 & 6.40 $\pm$ 0.74
& 18.53 $\pm$ 1.11 & 20.89 $\pm$ 1.43 & 5.17 $\pm$ 1.00 \\
&$Concat$    
& 19.24 $\pm$ 1.54 & 24.11 $\pm$ 2.07 & 6.91 $\pm$ 1.36
& 17.38 $\pm$ 0.46 & 21.36 $\pm$ 0.90 & 5.34 $\pm$ 0.35 \\
&$PVC$ \cite{li2014partial}   
& $\underline{26.93 \pm 0.40}$ & $\underline{31.40 \pm 0.73 }$& $\underline{12.60 \pm 0.20}$
& $\underline{24.80 \pm 1.05}$ & 28.04 $\pm$ 1.18 & $\underline{10.64 \pm 0.63}$ \\
&$MIC$ \cite{shao2015multiple}       
 & 22.95 $\pm$ 1.15 & 28.86 $\pm$ 1.09 & 9.41 $\pm$ 0.45
& 21.90 $\pm$ 1.27 & 25.79 $\pm$ 1.05 & 7.81 $\pm$ 0.73 \\
&$DAIMC$ \cite{hu2019doubly} 
 & 24.33 $\pm$ 1.30 & 29.25 $\pm$ 1.71 & 10.44 $\pm$ 1.02
& 22.84 $\pm$ 0.80 & 25.93 $\pm$ 1.17 & 8.47 $\pm$ 0.65 \\
&$Completer$ \cite{lin2021completer}        
& 25.32 $\pm$ 0.86 & 30.28 $\pm$ 0.61 & 10.32 $\pm$ 0.90
& 24.09 $\pm$ 0.97 & $\bm{31.24 \pm 1.58}$ & 7.35 $\pm$ 1.08 \\
&$DSIMVC$ \cite{tang2022deep}     
& 19.70 $\pm$ 0.76 & 21.39 $\pm$ 0.55 & 6.67 $\pm$ 0.23
& 18.87 $\pm$ 0.34 & 19.91 $\pm$ 0.55 & 6.01 $\pm$ 0.29 \\
&$DIMVC$  \cite{xu2022deep} 
& 24.27 $\pm$ 1.22 & 31.32 $\pm$ 1.06 & 11.56 $\pm$ 0.70
& 23.70 $\pm$ 1.50 & 29.14 $\pm$ 1.28 & 10.05 $\pm$ 1.30 \\
&$ICMVC (Ours)$ 
& $\bm{27.76 \pm 0.97}$ & $\bm{31.57 \pm 1.08}$ & $\bm{14.50 \pm 0.68}$
& $\bm{26.94 \pm 0.46}$ & $\underline{29.67 \pm 1.14}$ & $\bm{12.92 \pm 0.26}$\\ 
\hline
\multirow{9}{*}{Handwritten} 
&$BSV$  	
& 70.03 $\pm$ 5.33 & 69.99 $\pm$ 2.56 & 59.40 $\pm$ 4.42
& 66.01 $\pm$ 8.04 & 60.24 $\pm$ 4.72 & 46.60 $\pm$ 7.65
 \\
&$Concat$    
& 73.20 $\pm$ 8.43 & 71.82 $\pm$ 4.99 & $\underline{62.70 \pm 7.81}$
& 66.51 $\pm$ 4.92 & 63.15 $\pm$ 3.41 & 50.52 $\pm$ 4.02
 \\
&$PVC$ \cite{li2014partial}   
 & - & - & -
& - & - & -
\\
&$MIC$ \cite{shao2015multiple}       
& - & - & - & - & - & - \\
&$DAIMC$ \cite{hu2019doubly} 
& - & - & - & - & - & - \\
&$Completer$ \cite{lin2021completer}        
& $\underline{76.10 \pm 3.25}$ & $\underline{77.56 \pm 3.05}$ & 61.94 $\pm$ 6.96
& $\underline{76.61 \pm 4.89}$ & $\underline{77.84 \pm 1.54}$ & $\underline{63.16 \pm 6.04}$\\
&$DSIMVC$ \cite{tang2022deep}     
 & 74.79 $\pm$ 3.15 & 71.48 $\pm$ 2.16 & 61.95 $\pm$ 3.59
& 73.05 $\pm$ 3.81 & 69.64 $\pm$ 1.99 & 60.16 $\pm$ 3.28\\
&$DIMVC$  \cite{xu2022deep} 
& 61.81 $\pm$ 4.69 & 63.48 $\pm$ 3.96 & 48.45 $\pm$ 5.74
& 53.28 $\pm$ 4.84 & 50.29 $\pm$ 7.82 & 36.71 $\pm$ 7.92\\
&$ICMVC (Ours)$ 
& $\bm{84.95 \pm 1.31}$ & $\bm{83.62 \pm 1.72}$  & $\bm{78.01 \pm 2.28}$
& $\bm{82.34 \pm 1.62}$ & $\bm{80.24 \pm 1.58}$  & $\bm{73.22 \pm 2.41}$\\ 
\hline
\multirow{9}{*}{Noisy MNIST} 
&$BSV$  	
&54.40 $\pm$ 2.31   &48.49 $\pm$ 1.33   &37.12 $\pm$ 2.06
&51.96 $\pm$ 5.11	&46.98 $\pm$ 1.40	&35.07 $\pm$ 2.69
 \\
&$Concat$    
&44.56 $\pm$ 3.90	&46.64 $\pm$ 2.22	&31.83 $\pm$ 2.73	
&45.17 $\pm$ 4.38	&46.20 $\pm$ 1.55	&31.98 $\pm$ 2.71
 \\
&$PVC$ \cite{li2014partial}   
&40.83 $\pm$ 1.24	&35.50 $\pm$ 0.46	&23.00 $\pm$ 0.48		
&45.32 $\pm$ 2.30	&35.19 $\pm$ 1.29	&24.03 $\pm$ 1.15
\\
&$MIC$ \cite{shao2015multiple}       
&32.79 $\pm$ 1.65	&31.91 $\pm$ 0.74	&16.02 $\pm$ 0.68		
&29.26 $\pm$ 1.25	&28.84 $\pm$ 1.37	&12.76 $\pm$ 0.68
\\
&$DAIMC$ \cite{hu2019doubly} 
&39.68 $\pm$ 0.67	&37.11 $\pm$ 0.49	&24.95 $\pm$ 0.35		
&39.67 $\pm$ 1.94	&33.79 $\pm$ 1.37	&22.60 $\pm$ 1.40
\\
&$Completer$ \cite{lin2021completer}        
&81.82 $\pm$ 11.02  &$\underline{82.44 \pm 4.27}$   &74.76 $\pm$ 10.83		
&$\underline{79.96 \pm 2.23}$ &$\underline{80.08 \pm 1.13}$   &$\underline{73.57 \pm 2.52}$
\\
&$DSIMVC$ \cite{tang2022deep}     
&$\underline{85.59 \pm 8.27}$	&80.10 $\pm$ 7.40	&$\underline{76.51 \pm 10.38}$		
&73.83 $\pm$ 4.62	&67.25 $\pm$ 2.65	&60.06 $\pm$ 4.22
\\
&$DIMVC$  \cite{xu2022deep} 
&51.38 $\pm$ 12.03  &52.66 $\pm$ 13.89  &38.65 $\pm$ 13.13		
&57.30 $\pm$ 6.23	&55.89 $\pm$ 8.44	&44.11 $\pm$ 9.13
\\
&$ICMVC (Ours)$ 
&$\bm{97.94 \pm 0.05}$	&$\bm{94.57 \pm 0.12}$	&$\bm{95.51 \pm 0.11}$		
&$\bm{96.82 \pm 0.11}$	&$\bm{91.96 \pm 0.21}$	&$\bm{93.13 \pm 0.23}$
\\ 
\hline

\end{tabular}
}
\captionof{table}{Table of the clustering results on five datasets,  the $1^{st}$ and $2^{nd}$ best results are indicated in bold and underlined, respectively. $'-'$ means the result is unavailable due to non-negative constraint violation.} 
\label{table:evalutionX}
\end{table*}

Due to space limit, we only present the average values of the experimental results in the main body. Here, we supplement the full experimental results. Specifically, we run each dataset five times to calculate the standard deviation, and the  full experimental results are shown in Table~\ref{table:evalutionX}.

\begin{figure*}[!hbp]
\vspace{-5em}
\centering
	\begin{minipage}{0.245\textwidth}
		\includegraphics[height=3.5cm, width=1\linewidth]{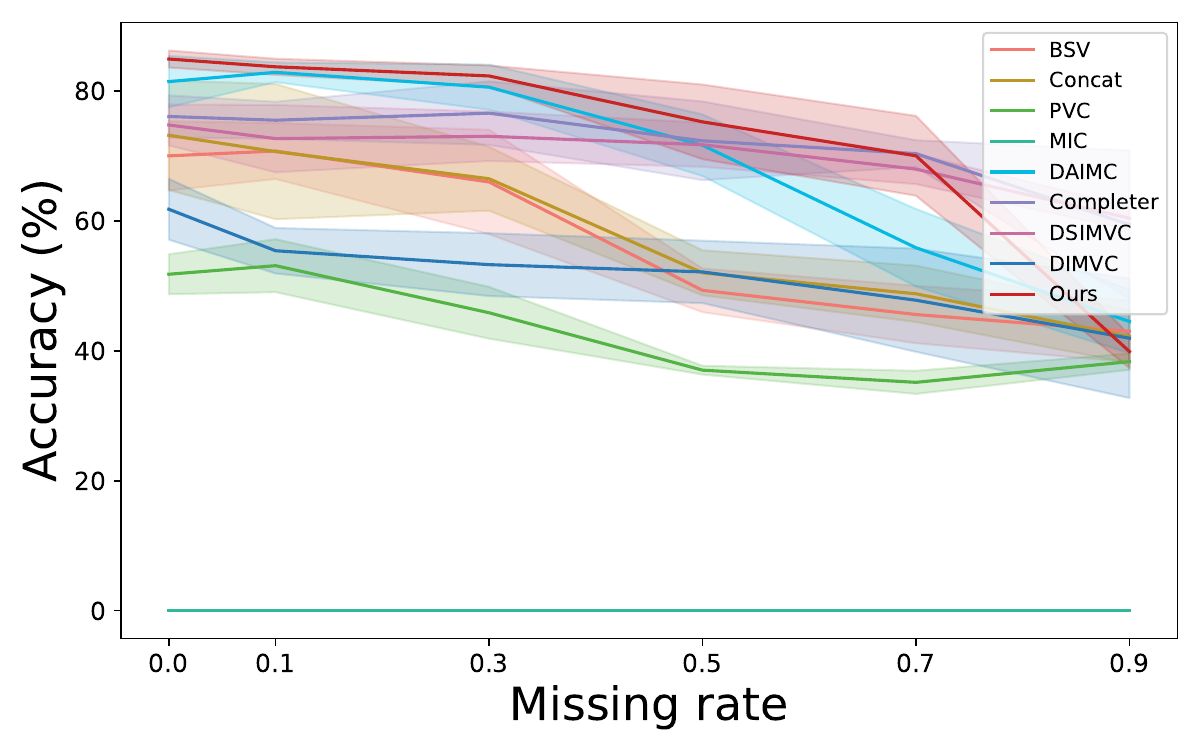}
	\end{minipage}
        \begin{minipage}{0.245\textwidth}
		\includegraphics[height=3.5cm, width=1\linewidth]{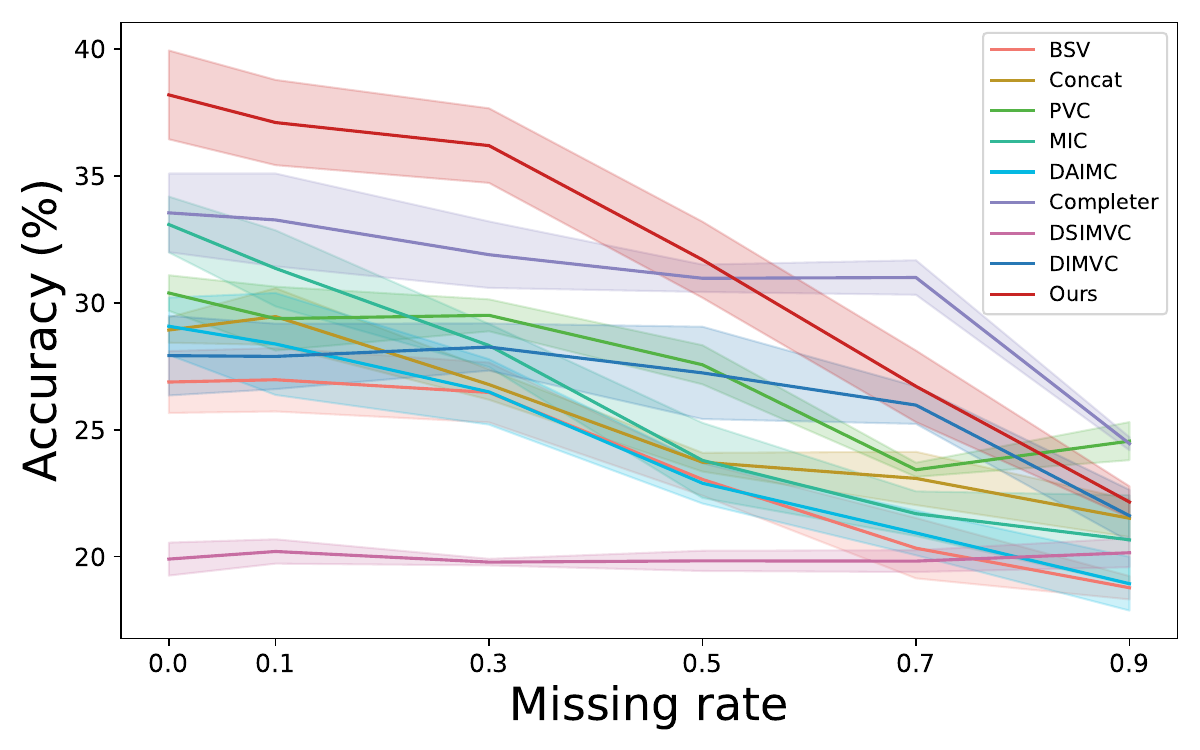}
	\end{minipage}
        \begin{minipage}{0.245\textwidth}
		\includegraphics[height=3.5cm, width=1\linewidth]{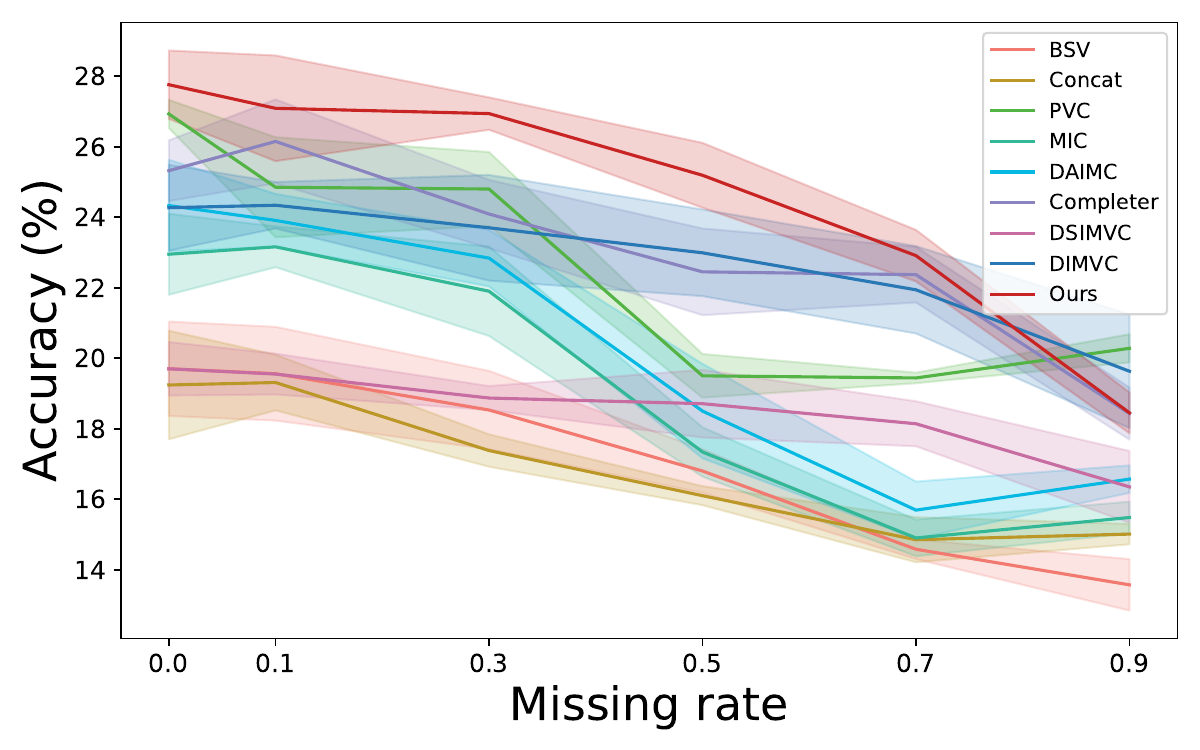}
	\end{minipage}
    \begin{minipage}{0.245\textwidth}
		\includegraphics[height=3.5cm, width=1\linewidth]{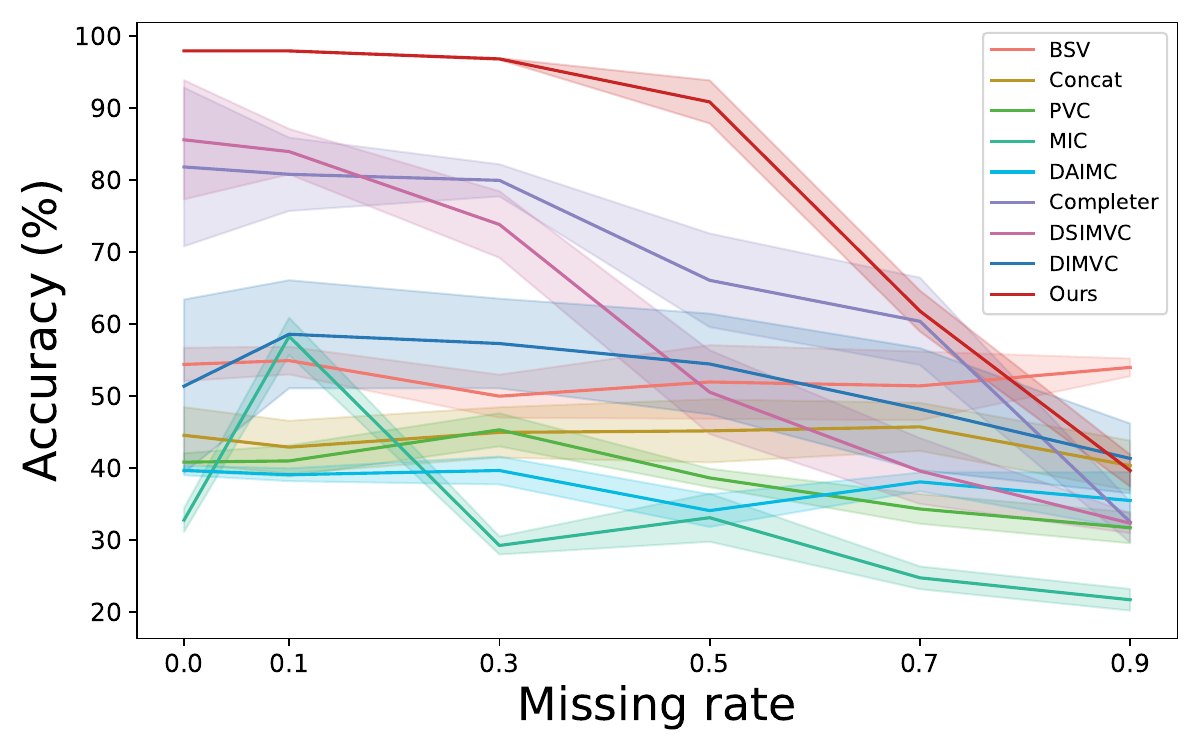}
	\end{minipage}
        \begin{tabularx}{0.95\linewidth}{ZZZZ}
        Handwittern & Scene-15 & LandUse-21 & NoisyMNIST
        \end{tabularx}
    \caption{Performance error band plots on other datasets as the missing rate increases.}
    \label{fig:exp-appen-miss}
\end{figure*}

In addition, as shown in Figure \ref{fig:exp-appen-miss}, we also show the clustering performance of our method at different missing rates on other datasets. It can be seen that our method performs well under low missing rates as described in the main body.

\end{document}